\documentclass[journal]{IEEEtran}

\usepackage{ifpdf}
\usepackage{authblk}

\usepackage{cite}

\ifCLASSINFOpdf
  \usepackage[pdftex]{graphicx}
\else
  \usepackage[dvips]{graphicx}
\fi

\usepackage{amsmath,bm}
\usepackage{amsfonts}
\usepackage{amssymb}
\usepackage{color,soul}
\usepackage{soul,color}
\soulregister\cite7
\soulregister\ref7

\ifCLASSOPTIONcompsoc
\usepackage[caption=false,font=normalsize,labelfont=sf,textfont=sf]{subfig}
\else
   \usepackage[caption=false,font=footnotesize]{subfig}
\fi

\usepackage{algorithmic}

\usepackage{array}

\usepackage{fixltx2e}

\newcommand\blfootnote[1]{%
  \begingroup
  \renewcommand\thefootnote{}\footnote{#1}%
  \addtocounter{footnote}{-1}%
  \endgroup
}

\begin{document}
\title{A CNN-LSTM Quantifier for Single Access Point CSI Indoor Localization}
\author{Minh Tu Hoang, Brosnan Yuen, Kai Ren, Xiaodai Dong, Tao Lu, Robert Westendorp, Kishore Reddy, and Hung Le Nguyen}
\maketitle

\begin{abstract}
\blfootnote{
This work was supported in part by the Natural Sciences and Engineering Research Council of Canada under Grant 520198, Fortinet Research under Contract 05484 and NVidia under GPU Grant program (\textit{Corresponding authors: X. Dong and T. Lu.}). \\
M. T. Hoang, B.Yuen, X. Dong and T. Lu are with the
Department of Electrical and Computer Engineering, University of Victoria,
Victoria, BC, Canada (email: \{xdong, taolu\}@ece.uvic.ca).\\
R. Westendorp and K. Reddy are with Fortinet Canada Inc., Burnaby, BC,
Canada.}

Earlier indoor localization algorithms used convolutional neural networks (CNNs) and memoryless models for estimating the positions and orientations of mobile devices. Unlike the aforementioned methods, we invented a hybrid CNN-LSTM model, leveraging both spatial and temporal features of channel state information (CSI) from a single router to precisely predict mobile device locations. Moreover, our CNN-LSTM goes beyond conventional limited classification methods by adopting a quantification approach that offers more accurate estimation, particularly for locations that are not in the training dataset. To address the instability inherent in CSI data, we introduce a comprehensive filtering and normalization scheme, which effectively mitigates this issue. Multiple extensive on-site experiments were used to evaluate the localization accuracy. Specifically, hundreds of testing positions were collected on the  Nexus 5 smartphone and the Intel 5300 NIC laptop across multiple days. Our structure indicates the average localization error of 2.5~m in a 21~m by 16~m test site with only a single WiFi router. Remarkably, our approach outperforms state-of-the-art algorithms by approximately $\mathrm{50\%}$ under identical test conditions. The results of our paper also reveal the accuracy limitation of using a single router to do localization. Our proposed method and database can be used for the comparison baseline for other research.  

\end{abstract}




\section{Introduction} \label{sec:intro} 

Indoor localization plays an indispensable role across various applications like autonomous navigation, virtual reality, contact tracing, and many others ~\cite{Xu2016}. Every new wireless device on the market now integrates the IEEE 802.11 standard, known as WiFi, onto its chipset. As a consequence, WiFi fingerprinting is now ubiquitous for tracking any kind of wireless device. Many WiFi fingerprinting localization systems rely on received signal strength indicator (RSSI) for predicting the pose of mobile devices, where the RSSI is conventionally measured at the WiFi receiver side of mobile phones, wireless headphones, portable speakers, tablets, and laptops ~\cite{Chen2016}. However, RSSI has significant flaws in its drastic signal fluctuations~\cite{Haochen2017} and insufficient information available for localization. For example, one single router provides only one RSSI reading in a frequency channel at one time. Therefore, in order to locate the accurate positions of users, a considerable number of access points (APs) is required. For this reason, 6 APs were used in SRL-KNN~\cite{Minh2018} and 15 APs in DANN~\cite{Fang2008}. The endless mall renovations and the ever changing shop layouts makes the very large numbers of APs for indoor localization  infeasible and expensive~\cite{Han2019}. Furthermore, in practical scenario, many small to medium area sites have only a single available AP such as in a small store, local pharmacy or classroom, which is impossible to use only RSSI for accurate localization. In order to address that problem, additional finger prints in needed.  As opposed to RSSI, channel state information (CSI) contains a set of fine-grained values from the physical layer (PHY) which include both amplitude and phase components of multiple subcarriers in frequency domain~\cite{Wu2013}. In contrast to having only one RSSI per packet, multiple CSI values corresponding to different subcarriers can be obtained from a single packet. The rich information in CSI enables localization with a single AP. For this reason, this paper adopts CSI and explores the algorithm to locate accurately a mobile device using only a single WiFi AP in a medium indoor environment area.  

\section{Related Works} \label{sec:related} 

Due to CSI having unique phases and amplitudes of each subcarrier, key information like the time of arrival (ToA) and angle of arrival (AoA) can be discerned from the CSI. FILA~\cite{Wu2013} gauges the signal strengths of WiFi packets by computing the received power of individual subcarriers at every possible location and AP. Afterwards, Bayes' rule is applied to the processed data in order to predict the positions of mobile devices. The experiment shows that FILA obtains a 40$\%$ reduction in localization error compared with RSS based Horus~\cite{Youssef2005}. Because the summation of the subcarriers' power in FILA~\cite{Wu2013} has the drawback of losing the diversity information, DeepFi~\cite{Wang2017} proposed a novel deep learning system directly utilizing the CSI amplitudes. In the training phase, the deep learning approach exploits the CSI amplitudes from multiple subcarriers to train the network weights and uses them as unique identifiers. In the testing phase, a probabilistic algorithm based on the radial basis function (RBF) is used to obtain the user's pose. With only one AP being utilized, DeepFi has the best accuracy of $\mathrm{0.94{\pm}0.56~m}$ which outperforms several existing RSS and CSI based schemes. Different from DeepFi, ConFi~\cite{Haochen2017} organizes the CSI amplitudes into a time-frequency matrix that resembles CSI images. These images are the fingerprint features for each location. ConFi models localization as a classification problem and addresses it with a five-layer CNN that consists of three convolutional layers and two fully connected (FC) layers. In~\cite{Minh2020}, the enhanced models of FILA and ConFi with an additional semi-sequential step are proposed to boost up their performance by around 25$\%$.

Beside the amplitude, CSI phase is also widely utilized for localization. For example, PhaseFi~\cite{Wang2016a} first extracts the raw phase value from the complex CSI and removes the offset through linear transformation to get the calibrated phase. A deep network with three hidden layers is adopted to train the calibrated phase data, and probabilistic Bayes is used for location estimation. Instead of using the coarse phase value directly as in PhaseFi, CIFI~\cite{Wang2018} extracts the phase of CSI to estimate the phase differences between two adjacent antennae or the angle of arrival (AoA). AoAs are constructed in a form of image to feed to deep convolutional neural networks (DCNN) for indoor localization. The best accuracy of these methods is around $\mathrm{1.0{\pm}0.4~m}$~\cite{Wang2016a}. In BiLoc~\cite{BiLoc2017}, Wang \textit{et al.} combine both average CSI amplitudes over pairs of antennae and estimated AoAs to form a bi-modal data. A complicated network that incorporates deep autoencoder, restricted Boltzmann machine along with radial basis function is utilized to estimate the user's position. The method is implemented in a total of 25 testing points in two different environments including a computer lab and a corridor, with a best accuracy of $\mathrm{1.5{\pm}0.8~m}$.   

Based on CSI, some research works can exploit ToA to locate the user's position by trilateration method~\cite{Vasisht2015,Ricciato2018}. In~\cite{Vasisht2015}, a set of algorithms named Chronos is proposed to estimate the ToA relying on the phase values after inverse fast Fourier transform (IFFT) of the original CSI. A hopping method between multiple frequency bands is utilized to increase the accuracy of estimated ToA. Based on the experimental results, Chronos achieves 0.47~ns median time-of-flight error, corresponding to a physical localization accuracy of 14.1~cm. Ref.~\cite{Ricciato2018} estimates ToA with the super-resolution algorithm~\cite{Li2004} by transforming the CSI to time domain pseudospectrum, and TOA being obtained by detecting the first peak of the pseudospectrum in the delay axis. The minimum resolution~\cite{Ricciato2018} is 10~ns, corresponding to a physical distance of 3~m.

\begin{table*}[h]
\centering         
\caption{Evaluations of various CSI indoor positioning methods } \label{table:ExpCompare} 
\begin{tabular}{l c c c c c c c c} 
\hline           
\textbf{Methods} & \textbf{Features} & \textbf{Access Points (APs)} & \textbf{Sample Locations (SLs)} & \textbf{Testing Points} & \textbf{Testing Selection} & \textbf{Accuracy}\\ 
FILA~\cite{Wu2013} &  Amplitude  &  1-3 &  28 &  -  & Fixed & 0.4 m to 1 m \\
DeepFi~\cite{Wang2017} & Amplitude  &  1 &  50 &  30 & Fixed & 0.94 $\pm$ 0.56 m\\
ConFi~\cite{Haochen2017} &  Amplitude  &  1 &  64 &  10  & Fixed & 1.3 $\pm$ 0.9 m\\
PhaseFi~\cite{Wang2016a} &  Phase  &  1 &  38 &  12 & Fixed & 1.0 $\pm$ 0.4 m\\
CIFI~\cite{Wang2018} & Phase  &  1 &  15 &  15 & Fixed  & 1.7 $\pm$ 1.2 m\\
BiLoc~\cite{BiLoc2017} & Amplitude $\&$ Phase & 1 & 25 & 25 & Fixed & 1.5 $\pm$ 0.8 m\\
\textbf{Proposed CNN-LSTM} & Amplitude & 1 & \textbf{1,185} & \textbf{195} & \textbf{Random} & See Table. \ref{table:AverageErr2}\\
\hline         
\end{tabular} 
\end{table*}

Up to date, only a limited number of IEEE 802.11 chipsets are able to provide CSI readings. These include Intel WiFi Link 5300 MIMO NIC, and devices built on Atheros AR9390 or AR9580 chipset~\cite{Wang2017b}.  Recently, Schulz \cite{Matthias2018} have developed Nexmon, a firmware patching framework for Nexus 5 smartphone. Nexmon can extract the CSI data from the PHY and send them to the user's interface through UDP frames.     

Despite of extensive investigation, some of the following issues still exist in all of the above methods. 
\begin{itemize}
\item[1] Previous CSI experiments were limited by the insufficient number of sample locations (SLs) and testing points per unit testing area, as the measurements on the SLs are conducted manually. Table~\ref{table:ExpCompare} compares the experimental areas and results between different algorithms. The largest number of SLs is 64 in ConFi~\cite{Haochen2017}, and that of testing points is 30 in DeepFi~\cite{Wang2017}. As the localization accuracy is determined by the density of the SLs in the target area, the performances of the reported algorithms are limited.
\item[2] CSI fluctuations are essential to localization accuracy, but rarely investigated in past research. With the presence of a human, CSI readings from a subchannel can decrease (or increase) significantly~\cite{Shi2018}. As a result, CSI fingerprints in the database may not match the instant CSI readings in the testing phase. 
\item[3] The information of the previous time steps in user's trajectory has not been exploited to enhance localization. 

Assuming human walking velocities are significantly slower than WiFi packet update rates, the historical data from previous steps can provide useful information to predict the current user's location.              
\end{itemize}  

This paper proposes a single AP indoor localization algorithm, and introduces the solution to the problems listed above for mobile devices, including phones and laptops. The main contributions of this paper are

\begin{itemize}
\item[1] For the first time according to authors' knowledge, extensive autonomous experiments are conducted to collect WiFi CSI data in both phone (Nexus 5) and laptop (Intel WiFi 5300 NIC card installed in a Thinkpad laptop) at thousands of SLs and along several testing trajectories in various time of the day. Since the automated procedure enables us to acquire a large number of data, we are no longer restricted to have the testing point position to be at the same location of an SL or at a fixed position that is a priori known. In our experiments, the testing points are selected anywhere in the target area and estimated through a quantification model. This results in higher localization accuracy in practice than the traditional classification  models. With the sufficient amount of data, we analyze the temporal changes of CSI and propose more comprehensive solutions to reduce the inaccuracy from CSI fluctuation.
\item[2] This paper compares the performance of our work to previously reported single AP localization algorithms in Table~\ref{table:ExpCompare} and Table~\ref{table:AverageErr2}. It is worth mentioning that previous work assume the testing point locations are a prior known and apply classification for localization. Consequently, localization accuracy is limited by the SL density and their performance in real scene is further reduced. In contrast, as a large number of measurements can be done with an autonomous navigating robot, our quantification based localization model does not require the testing point a prior known and its generalization to real scene will not reduce the performance. The results of our paper also revealed the accuracy limitation of using a single router to do localization. Our proposed method and database can be used for the comparison baseline for other research. 
\item[3] For the first time in single AP localization, this paper combines convolutional neural network (CNN) for CSI feature extraction and long-short term memory network (LSTM) to impart temporal correlation of data to achieve high accuracy even in practical scenarios. 
\end{itemize} 

\section{Proposed Method} \label{sec:proposed} 

\subsection{Localization System Overview}
The indoor positioning system has two distinct phases: the offline training phase and the online testing phase. For the training phase, the CSI at every preplanned sample location (SL) is recorded for training the machine learning models. Since only one single AP is used along with $M$ collected SLs, each SL $i$ at its physical location $\bm{l}_{i}(x_{i},y_{i})$ has a corresponding CSI from multiple antennae and multiple subcarriers. Each individual CSI value contains a real and an imaginary component that encodes phase and amplitude.  Here, CSI amplitude, phase, ToA or any combination of them are formed in the shape of an image for CNN feature extraction. Since CSI phase is prone to noise  due to fading and frequency offset~\cite{Haochen2017,Wang2017}, complicated preprocessing is needed before using it as a feature~\cite{Haochen2017}. In addition, it is difficult to obtain accurate ToA that requires line of sight (LOS). To simplify the computation and avoid heavy preprocessing procedure, we choose CSI amplitude as the fingerprint for our proposed method. We assign all data points belonging to the same subcarrier into the same column and place all data points measured at the same time into the same row to construct a matrix
\begin{equation} \label{eq:am_csi}
\centering   
{\tilde A}(\bm{l}_{i}) = 
\begin{bmatrix}
    {A}^{11}_i       & {A}^{12}_i  & \ldots & {A}^{1W}_i \\
    {A}^{21}_i       & {A}^{22}_i  & \ldots & {A}^{2W}_i \\
    \vdots          & \vdots    & \ddots & \vdots \\
    {A}^{H1}_i       & {A}^{H2}_i  & \ldots & {A}^{HW}_i \\
\end{bmatrix}.
\end{equation}
where, ${A}^{hw}_i$ is the h-th measurement of w-th subcarrier at SL $i$. ${\tilde A}(\bm{l}_{i})$ is then treated as a $H{\times}W$ ``image'' that contains WiFi channel information of location $i$. In the rest of the paper, we adopt deep learning based image processing techniques for indoor localization.  During the training phase, a self navigating robot will stop by each SL to collect a large number  ($W_{1}$) of CSI images. In the testing phase, as a user is expect to move from time to time, only a small number of CSI images ($W_{2}$) will be collected at each testing point. 

\subsection{CNN-LSTM Overview}
CNN is a deep learning architecture that is known to be capable of extracting features from images for classification and recognition~\cite{Cun1998,Simard2003}. In the realm of CNNs, three crucial layers form the backbone of image processing and analysis. These layers are the convolutional layer, max pooling layer, and fully connected layer, each playing a vital role in extracting insights from visual data. Specifically, the convolutional layer extracts valuable insights from images by pinpointing key spatial patterns, while the max pooling layer eliminates unnecessary data, reducing computational overhead and boosting efficiency. Finally, the fully connected layer can convert the multi-dimensional output of the convolutional layer to a one-dimension vector to be effectively processed in the following layer of the network.

Unlike Convolutional Neural Networks (CNNs), recurrent Neural Networks (RNNs) possess a unique strength, internal memory storage, enabling them to recall and leverage historical data, such as past RF signals ~\cite{Ho2017}. This capacity makes RNNs ideal for applications where data exhibits sequential correlations. In indoor localization, for instance, a user's current location is intimately tied to their previous positions, as movement is limited within a given time frame. By harnessing the sequential patterns in Channel State Information (CSI) and Received Signal Strength Indication (RSSI) measurements along the user's trajectory, RNNs can significantly enhance localization accuracy. Up to date, there are several RNN variants. Among them, the vanilla RNN~\cite{Zachary2015} has a few performance flaws because the gradient rapidly approaches zero when stacking multiple RNN layers~\cite{Chung2014}. To address this limitation, long short-term memory (LSTM)~\cite{Hochreiter1997} introduces an internal memory mechanism, enabled by a forget gate, which regulates the influence of previous inputs and temporal dependencies. Further, gated recurrent unit (GRU)~\cite{Cho2014} reduces forget, update and output gates from LSTM to update and reset gates only. Bidirectional RNN (BiRNN)~\cite{Zachary2015}, bidirectional LSTM (BiLSTM)~\cite{Schuster1997} and bidirectional GRU (BiGRU)~\cite{Zhao2018} extend traditional RNN, LSTM and GRU respectively by utilizing all available input information from not only the past but also the future of a specific time frame. In \cite{Minh2019}, Hoang \textit{et al.} evaluated various RNN architectures, including vanilla RNN, LSTM, and their bidirectional counterparts, for indoor localization tasks. Through rigorous experiments utilizing WiFi RSSI, they determined that LSTM achieves superior accuracy compared to other models. Consequently, we adopt LSTM as our sequential model in this paper. The subsequent section will delve into the details of our proposed WiFi CSI indoor localization system, which leverages a hybrid CNN-LSTM approach.

\subsection{Proposed Localization System} \label{sec:proposed_system} 
\begin{figure*}[!t]
\centering
\includegraphics[width=\textwidth]{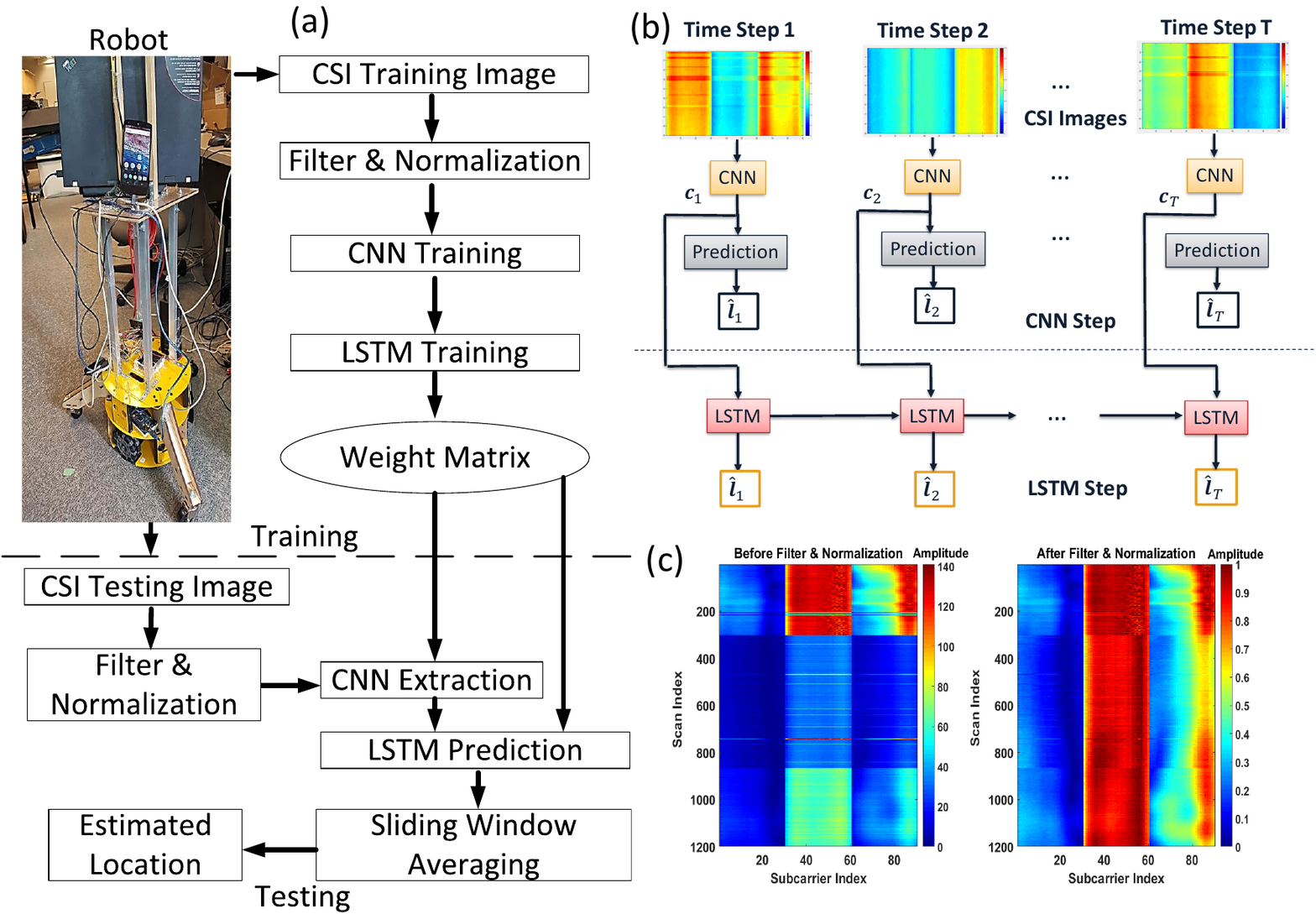}
\caption{(a) Localization process of the proposed CNN-LSTM system. (b) Proposed CNN-LSTM model. (c) CSI images before and after applying  filter and normalization.}
\label{fig:Loc_System}
\end{figure*}

The high level synopsis of the proposed localization system is presented in Fig.~\ref{fig:Loc_System}(a) with details specific to CNN-LSTM in Fig.~\ref{fig:Loc_System}(b). CSI data for both training and testing purposes is gathered by an autonomous robot depicted in the figure. The localization process is as follows.

\subsubsection{CSI Filter And Normalization}  \label{sub:data_filter} 
Though the modified firmware for Intel WiFi 5300 NIC card~\cite{Halperin2011} and Nexus 5 smartphone~\cite{Matthias2018}, the mobile device receives one CSI reading per beacon frame or data frame. While CSI values offer a wealth of information from multiple subcarriers, they are susceptible to environmental fluctuations, including human obstruction and movement, interference from nearby devices and equipment, and variations in receiver antenna orientation, among other factors~\cite{Shi2018}. In order to filter out the outliers caused by the interference and mitigate the signal instability, we adopt the median filter~\cite{Broesch2008} for each subcarrier followed by a normalization procedure to get cleaner data from the original image. The normalization is done by first average the amplitudes of each filtered image at each location $\bm{l}_{i}$ according to                                                                     
\begin{equation} \label{eq:average_am}
{A}_{i} =  \frac{\sum_{h=1}^{h=H} \sum_{w=1}^{w=W} {A}^{hw}_i}{H{\cdot}W}.
\end{equation} 
Denote $A_{max}$ as the largest average amplitude of $\{A_i\}$ and  $\bm{l}_{max}$ being the corresponding location, every CSI reading at every location (row of the CSI image) can then be normalized by the min-max normalization. To do so, define $A^{h}_{i,min}$ and $A^{h}_{i,max}$ as the minimum and maximum value of the $h$-th row in ${\tilde A}(\bm{l}_i)$. We can construct a normalized CSI image matrix ${\hat {\tilde A}}({\bm{l}_i})$ whose $(h,w)$ element ${\hat A}_i^{hw}$ is given by     
\begin{equation} \label{eq:min_max_norm}
{\hat A}^{hw}_i = \frac{A^{hw}_i - A^{h}_{i,min}}{A^{h}_{i,max} -A^{h}_{i,min}}\frac{{A}_i}{A_{max}}.
\end{equation} 
Here, $\frac{{A}_i}{A_{max}}$ is to preserve the original power of the CSI image.

Fig.~\ref{fig:Loc_System}(c) illustrates an example of CSI images collected by Intel WiFi 5300 NIC card before and after applying the filter and normalization process. There are noisy readings (scattered rows), caused by the fluctuation of the CSI amplitude, inside the original CSI image. The CSI image is constructed with 1200 scans/rows and 90 subcarriers/columns from 3 antennae. Before normalization, the amplitude of subcarriers in the CSI image is unstable among different scans. For example, the amplitudes of subcarriers 30 to 60 reach to as high as 140 (red color) in the first 300 scans and rapidly reduce to below 60 (light blue color) from scan 300 to 1200. In contrast, after applying the median filter and normalization process, the amplitude of the CSI image at the same location is more consistent as show in the figure.     
         
\subsubsection{Proposed CNN-LSTM Model} \label{sub:rnn_model}                                
The proposed CNN-LSTM model is trained on CSI images from successive locations in a trajectory, leveraging the temporal correlation between them. Each location in a trajectory corresponds to a distinct time instance. The length of a trajectory, or equivalently, the number of time steps, defines the memory length $T$ as illustrated in Fig.~\ref{fig:Loc_System}(b). Since all weights and hidden state values are stored at each time step in a training trajectory~\cite{Zachary2015}, the number of time steps $T$ influences the efficacy of LSTM and CNN-LSTM models. Larger $T$ incorporates more historical information but also exacerbates the temporal bias in predictions \cite{Minh2019}. Given that the distance between adjacent locations is restricted by the maximum distance a user can traverse within the sampling period, the CNN-LSTM training trajectories are randomly generated based on the method proposed in~\cite{Minh2019}.

There are two main layers in our CNN-LSTM model (Fig. \ref{fig:Loc_System}(b)) including CNN and LSTM layers. While the input of CNN layer is the CSI image from locations in a trajectory, the input of LSTM layer is the output of the previous CNN layer. In other words, CNN will extract the CSI information from the spatial domain, whereas LSTM will further exploit that information in the temporal domain. With the information from both domains, the ambiguity of CSI data is significantly reduced. Subsection~\ref{sec:sim_result} will show experimental results to demonstrate this reduction.

The objective of CNN-LSTM training is to minimize the loss function $\mathcal{L}(\bm{\hat{l}},\tilde{\bm{l}})$ which penalizes the Euclidean distance between the output $\bm{\hat{l}}$ and the target $\tilde{\bm{l}}$. To achieve this goal, the backpropagation algorithm uses the chain rule to calculate the derivative of the loss function $\mathcal{L}$ and adjusts the network weights by gradient descent~\cite{Zachary2015}. In \cite{Valente2019}, Valente \textit{et al.} show that CNN-LSTM model should be trained separately, instead of using only one common output and one loss function. Therefore, the training of our proposed network is divided into two steps. In step 1, the CNN layer is trained as a quantification model using CSI images from the database and the loss function  mean square error (MSE) described as        
\begin{equation} \label{eq:loss1}
 \mathcal{L}(\bm{l},\tilde{\bm{l}}) = ||\bm{l} - \tilde{\bm{l}}||_2.
\end{equation} 
The output of the final FC layers in CNN ($\bm{c}_T$) is extracted and adopted for LSTM training (Fig.~\ref{fig:Loc_System}(b)) in step 2. The loss function of this phase is still MSE, but for $T$ time steps following MIMO-LSTM model~\cite{Minh2019} as  
\begin{equation} \label{eq:loss2}
\mathcal{L}(\bm{\hat{l}},\tilde{\bm{l}}) = \frac{\sum_{i=1}^{T} ||\bm{\hat{l}}_i - \tilde{\bm{l}}_i||_2}{T}.
\end{equation}

\section{Database And Experiments} \label{sec:experiment}

\begin{figure*}[!t]
\centering
\includegraphics[width=\textwidth]{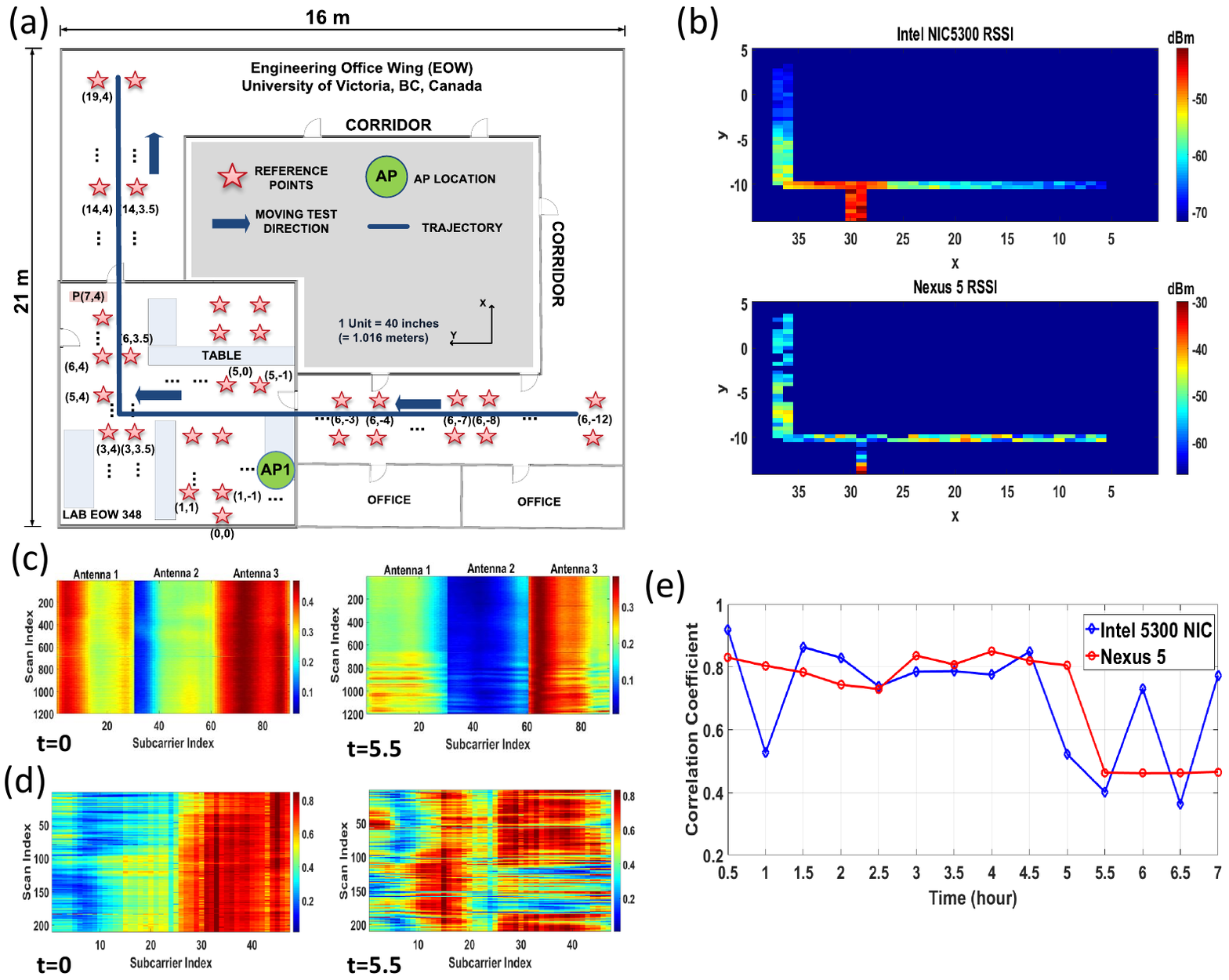}
\caption{(a) Diagram of the data collection points. The solid blue arrows indicate the walking direction of the testing route. (b) Heat map of AP RSSI signal collected from Intel 5300 NIC and Nexus 5 phone. (c) CSI waterfall plot from Intel 5300 PCIe card at location (0,0) in 2 different time. (d) CSI waterfall plot from Nexus 5 phone at location (0,0) in 2 different time. (e) Correlation coefficient of the collected CSI images at location (0,0) along 7 hours.}
\label{fig:floor_map}
\end{figure*}

The data collections took place on the 3rd floor of the Engineering Office Wing (EOW),  University of Victoria in British Columbia, Canada. As depicted in Fig.~\ref{fig:floor_map}(a), the office space spans twenty-one meters by sixteen meters, where it has multiple hallways and exits. We utilized and an autonomous robot (Fig.~\ref{fig:Loc_System}(a)) for collecting the CSI training and testing dataset, of which has a positional accuracy of $\mathrm{0.21{\pm}0.02~m}$  due to precise light detection and ranging (LIDAR), inertial measurement unit (IMU), RGB-D camera, and wheel encoders.

For the training and testing runs, the robot carried a Nexus 5 smartphone or a Thinkpad T520 laptop with an Intel 5300 PCIe card to collect CSI. There is only 1 AP (TPlink AC1750) in the experimental area, operating in channel 36 with a 5~GHz frequency band. Fig.~\ref{fig:floor_map}b depicts the heat map of the RSSI collected from the Intel 5300 NIC and Nexus 5 smartphone, where the signal strength is represented by colour. Visibly, the signals are stronger in the areas near the AP. The targeted AP can cover the area including one room (EOW 348) and two corridors with 1185~SLs (pink stars) and 195~testing points (solid blue line). There are 3 receiving antennae on the Intel 5300 NIC, while the smartphone has only one receiving antenna. Therefore, as shown in Fig.~\ref{fig:floor_map}(b), the RSSI signals from the Intel card are stronger and more consistent than the ones from the smartphone.  

In the training phase, the robot stays at every SL to collect data for 2 minutes. The CSI data is repeatedly collected in different days and at different time to build the database. For the testing phase, the robot navigates along predetermined waypoints shown in Fig.~\ref{fig:floor_map}(a) with velocities changing from 0.6 m/s to 4.0 m/s in order to imitate human walking behaviours~\cite{Browning2006,Email2007}. There are 195~testing points in each testing trajectory. The test experiment is conducted several rounds per day and repeated in 3 different days. Day 1 test mainly targets the quiet time when the area has fewer people (before 9 am and after 5pm). In contrast, during day 2 and 3 we test the localization accuracy at steady time (working hours) and busy time when more people moving around. Note that the validation data is collected at different time from the training data but at the same SL locations, while testing point locations are randomly selected to be different from all SLs in order to approximate our model closer to practical situations. The user location is updated every $\Delta_{t}=$1-2~s.   

\section{Results And Discussions} \label{sec:sim_result}
\subsection{CSI Sensitivity} \label{sub:Sensitivity}
The collected CSI data of each antenna on Intel 5300 NIC include the IEEE 802.11n channel matrices for 30 subcarrier groups, which is about one group for every two subcarriers at 20~MHz bandwidth~\cite{Halperin2011}. Fig.~\ref{fig:floor_map}(c) shows examples of CSI image in SL location (0,0) from this NIC with 1200 scans and 30 subcarriers for each antenna (total 90 subcarriers). Using modified Nexmon firmware~\cite{Matthias2018}, we obtain 64 IEEE 802.11ac-subcarriers at bandwidth 20~MHz with 47 of them providing non-zero CSI data from the single receiving antenna on the Nexus 5 phone. Fig.~\ref{fig:floor_map}(d) illustrates the CSI images from the phone in the same location (0,0) with 200 scans and 47 non-zero subcarriers.  

As mentioned in Subsection~\ref{sub:data_filter}, although CSI provides rich information from multiple subcarriers, it is sensitive to the change of environments, especially the interference of human blocking and movements. With the presence of a human, CSI measurements from a subchannel can vary widely~\cite{Shi2018}. Fig.~\ref{fig:floor_map}(e) illustrates the fluctuation of CSI data collected from both Intel 5300 NIC and Nexus 5 phone in SL location (0,0) over 7 hours. The experiment was conducted in working hours when many students (up to ${\sim}$10) used WiFi and moved around the area from time to time. CSI images of both devices were sampled every half an hour and were compared with the original CSI images collected at the beginning ($t=0$ h) using Pearson coefficient \cite{Minh2018}. Fig. \ref{fig:floor_map}(e) shows that CSI data of both devices are stable for a long period of time from $t=1.5$ h to $t=4.5$ h, with high correlation coefficients (${\sim}80\%$). It is the period when most students in the testing area were sitting in front of their desks. However, during the recess time, when people were moving around, the correlation coefficients of CSI data changed significantly with the drop to $40\%$ after 5 hours. Fig.~\ref{fig:floor_map}(c) and (d) illustrate the visualization of CSI images at $t=0$ h and $t=5.5$ h. We observe significant change of the image patterns in both Intel 5300 NIC and Nexus 5. Therefore, CSI fingerprints in the database might not match with the instant CSI readings in the testing phase because they were collected at different time.

According to our knowledge, this poses a significant challenge in localization but has been neglected in past research. Here, we for the first time propose a feasible solution using an autonomous robot to do the extensive experiments in different time slots. With the sufficient amount of collected training data, the mismatch between database and testing data is significantly reduced.   

\subsection{CNN Layer Analysis}
\begin{table}[!t]
\centering         
\caption{CNN Layer Parameters} \label{table:CNN} 
\begin{tabular}{c c c}          
\hline
\textbf{Category} & \textbf{Intel NIC}  & \textbf{Nexus 5}\\ 
Training Data & 47,400 CSI images & 35,000 CSI images \\
Validation Data & 23,700 CSI images & 14,000 CSI images \\
Conv Layer 1  &  Input($30 \times 30 \times 3$) & Input($10 \times 47 $)\\ 
    & $5 \times 5$ kernels, 10 filters &  $5 \times 5$ kernels, 10 filters \\
Conv Layer 2 & Input($30 \times 30 \times 10$) & Input($10 \times 47 \times 10$)\\ 
 & $5 \times 5$ kernels, 10 filters &  $5 \times 5$ kernels, 10 filters \\
Conv Layer 2 & Input($30 \times 30 \times 10$) & Input($10 \times 47 \times 10$)\\ 
 & $5 \times 5$ kernels, 10 filters &  $5 \times 5$ kernels, 10 filters \\
FC Layer 1  & 9000 neurons &  4700 neurons  \\
FC Layer 2   & 900 neurons &   470 neurons \\
Training Output  & Location $\tilde{\bm{l}}$(x,y) & Location $\tilde{\bm{l}}$(x,y) \\ 
Loss Function & MSE & MSE \\
 \hline  
\end{tabular} 
\end{table}

\begin{figure}[!t]
     \centering
\subfloat[\label{fig:CNN_Intel}]
{\includegraphics[width=0.5\textwidth]{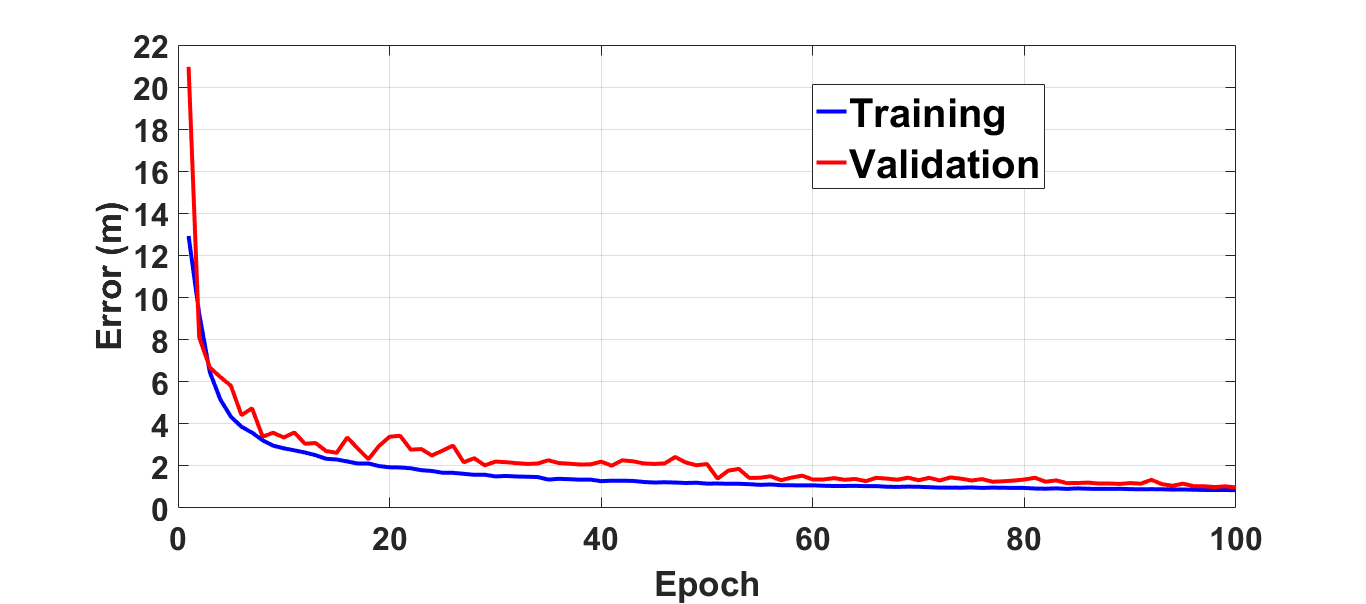}} \quad
\subfloat[\label{fig:CNN_Nexus}]
{\includegraphics[width=0.5\textwidth]{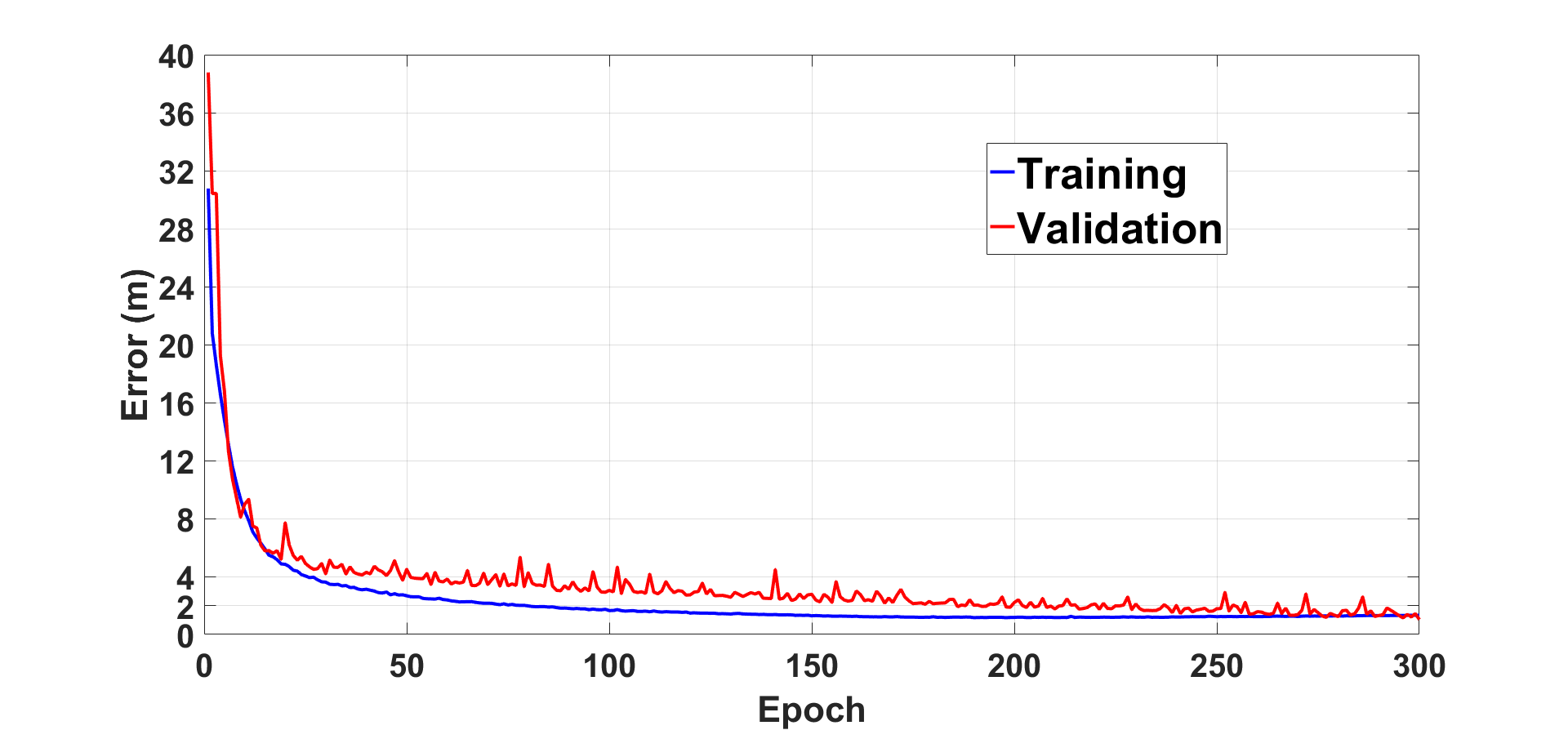}}
     \caption{CNN Layer Learning Curve (a) Intel 5300 NIC. (b) Nexus 5 Phone.}
     \label{fig:CNN_Curve}
\end{figure}

As explained in Section~\ref{sec:experiment}, the user's location will be updated every $\Delta_{t}=$1-2~s. Therefore, the CSI data are collected in the same time interval for CNN layer training and validation. During $\Delta_{t}$, the NIC receives an average of $30$ packets, and each CSI packet has total 90~subcarriers from 3 receiving antennae. We consider the data from each individual antenna as a channel of the CSI image, which creates the testing CSI image size for the NIC as ($30{\times}30{\times 3}$). On the other hand, the smartphone only receives $10$ scans during $\Delta_{t}$ with only one single receiving antenna and $47$ subcarriers. Therefore, the CNN training CSI image size for Nexus 5 is ($10{\times}47$).    

The CNN model is constructed based on ConFi~\cite{Haochen2017} with 3 convolutional layers (Conv) and 2 FC layers. Table~\ref{table:CNN} shows the detailed parameters of the proposed CNN structure for both NIC and smartphone. The differences between them are the inputs of each layer and the number of neurons for the FC layers. There are 9,000 and 900 neurons respectively in the two  FC layers for the Intel NIC, while 4,700 and 470 neurons in the case of using the Nexus 5 smartphone. 

In total, there are 47,400 NIC images for the NIC and 35,000 smartphone CSI images for CNN training and 23,700 and 14,000 images respectively for the validation. Fig.~\ref{fig:CNN_Curve}(a) and~\ref{fig:CNN_Curve}(b) illustrate the learning curves of CNN training for the NIC and smartphone data. Both training processes converge at the localization error around 1.7~m. On average, it takes 100 epochs for Intel 5300 NIC data training and 300 epochs for the smartphone. After training, CNN extracts spatial features, $\{\bm{c}_1, \bm{c}_2,\ldots,\bm{c}_T \}$, of $T$ CSI images from $T$ time steps from the output of the FC layer 2. In our experiment, we extract 900 output spatial features from every NIC CSI image, and 470 features for the smartphone.     
\begin{figure}[!t]
\centering
\includegraphics[width=0.52\textwidth]{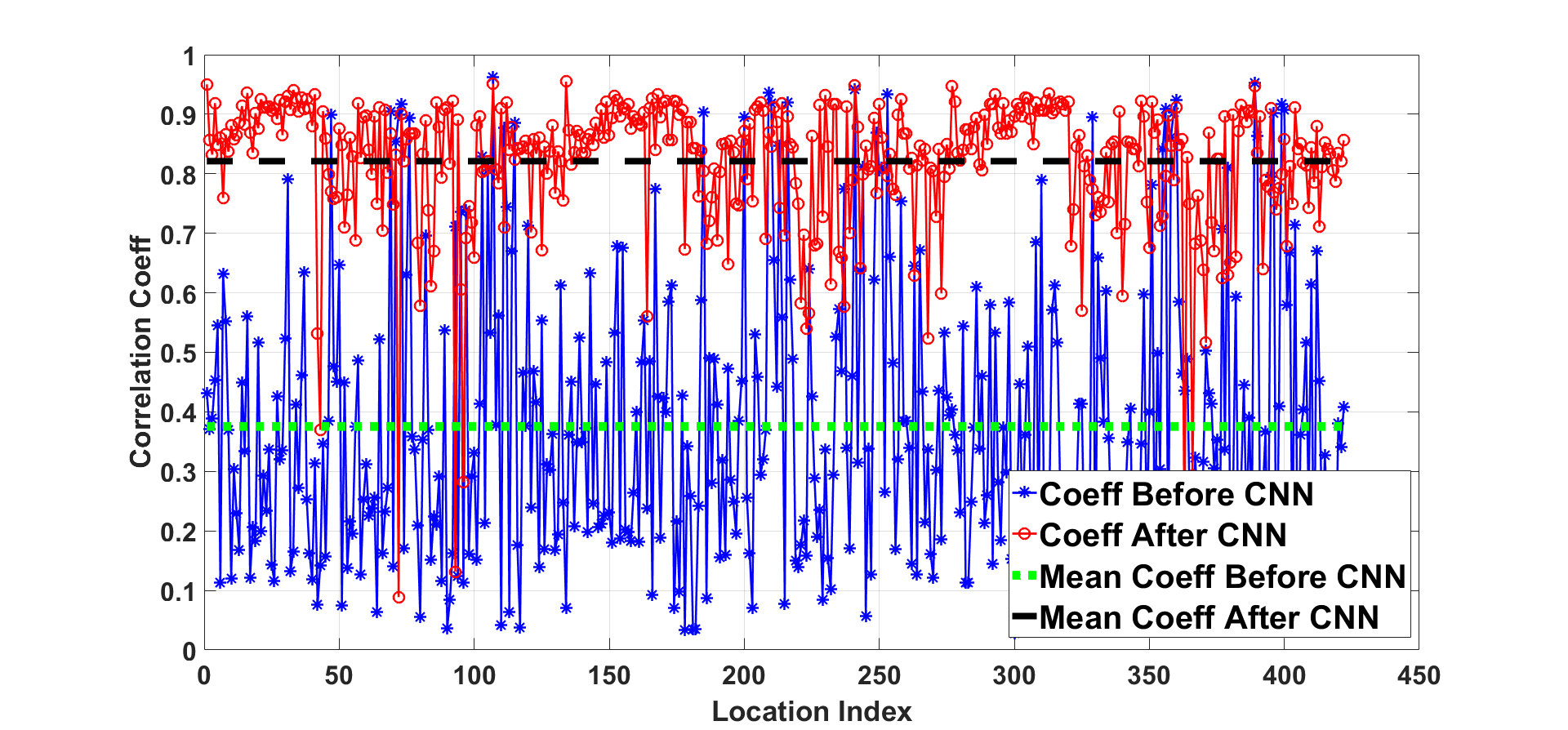}
\caption{Correlation coefficient of original CSI images before CNN and output spatial features after CNN with Intel 5300 NIC dataset.}
\label{fig:CNN_Corr}
\end{figure}

Fig.~\ref{fig:CNN_Corr} shows the significant increase of the CSI data correlation in 422~random SL locations after CNN layer. Each SL includes 120 CSI images collected at different time. The average correlation coefficient $\rho({\bm l_i})$ at each location ${\bm l_i}$ is defined as
\begin{equation}
\rho({\bm l_i})=\frac{\sum_{j=1}^N\sum_{k=1}^N \rho_{jk}({\bm l_i})}{N^2}.
\end{equation}
where $N=120$ is the number of CSI images at location ${\bm l_i}$, $\rho_{jk}({\bm l_i})$ is the Pearson coefficient between the $j^{th}$ image and $k^{th}$ image of that location. Before CNN, due to the CSI sensitivity explained in Subsection \ref{sub:Sensitivity}, the correlation between the original CSI images at different time but the same location is as low as around $40\%$. However, after CNN, the important spatial features are extracted from the original images with the irrelevant information being removed. Therefore, the correlation between output spatial features increases to around $80\%$. Similar results are obtained from the smartphone database. The CNN output spatial features, $\{\bm{c}_1, \bm{c}_2,\ldots,\bm{c}_T \}$, of $T$ will be fed to the next LSTM layer for further processing in time domain.   

\subsection{LSTM Layer Analysis}
\begin{table}
\centering         
\caption{Initial setup parameters for RNN system} \label{table:LSTM_layer} 
\begin{tabular}{c c c}          
\hline
\textbf{Category} & \textbf{Intel NIC} & \textbf{Nexus 5} \\ 
Memory length ($T$) & 5 & 5\\
Model & MIMO-LSTM & MIMO-LSTM \\
Loss function & MSE & MSE \\ 
Hidden layer & 900 neurons & 470 neurons \\
Dropout & 0.2 & 0.2\\
Optimizer & Adam & Adam \\
Learning rate & 0.001 & 0.001\\
Training Data & 30,000 trajectories & 30,000 trajectories \\
Validation Data & 15,000 trajectories & 15,000 trajectories\\
$\sigma$ & 2m & 2m\\
$\Delta_t$ & 1s & 1s\\
 \hline  
\end{tabular} 
\end{table}

\begin{figure}[!t]
     \centering
\subfloat[\label{fig:LSTM_Intel}]
{\includegraphics[width=0.5\textwidth]{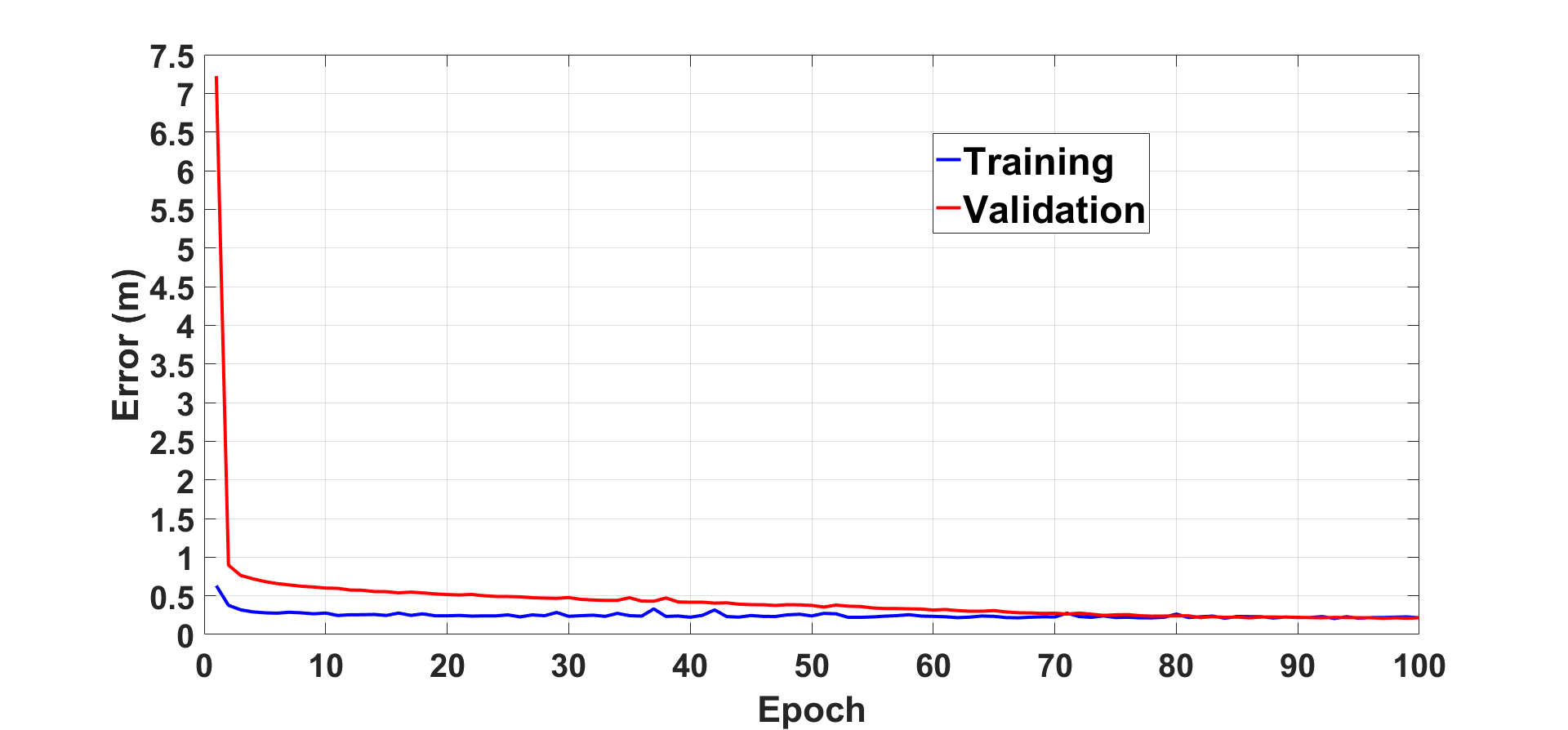}} \quad
\subfloat[\label{fig:LSTM_Nexus}]
{\includegraphics[width=0.5\textwidth]{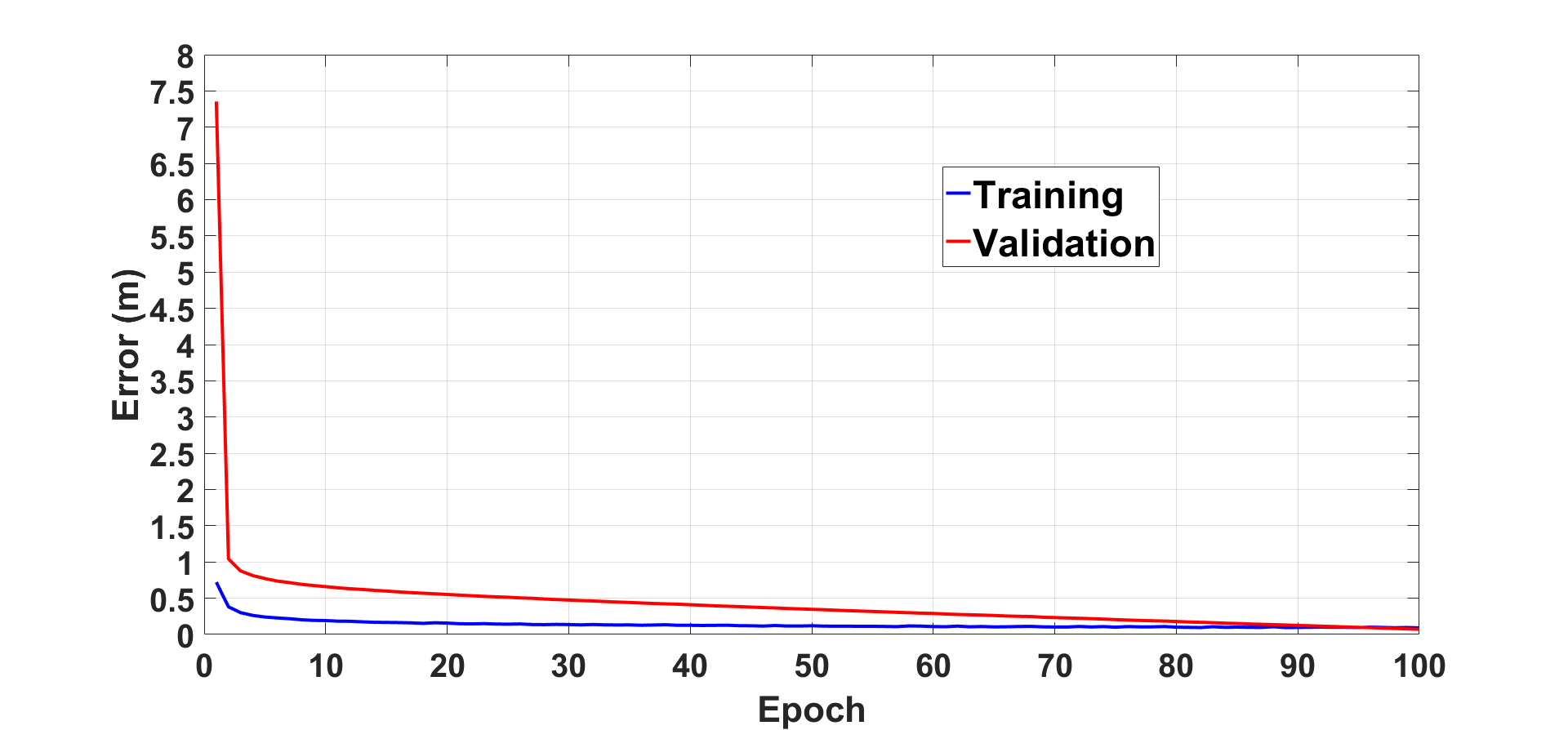}}
     \caption{LSTM Layer Learning Curve (a) Intel 5300 NIC. (b) Nexus 5 Phone.}
     \label{fig:LSTM_Curve}
\end{figure}

The detailed parameters for our LSTM network is shown in Table~\ref{table:LSTM_layer}. The LSTM model follows MIMO-LSTM proposed in~\cite{Minh2019}. The inputs are multiple generated trajectories containing CNN spatial features from $T$ time steps. Based on the trajectory generation method~\cite{Minh2019}, 30,000 training and 15,000 validation trajectories are generated randomly subject to the condition that the distance between successive locations is limited by the maximum distance ($\sigma$) a user can cover  within the sample interval $\Delta_t = 1$~s. In the conducted experiments in~\cite{Minh2019}, the authors show that the memory length $T=5$ and the other longer memory cases including $T=10$ and $T=40$ provide comparable results. Therefore, the memory length here is chosen as $T=5$. The outputs include $T$ time steps locations $\{\bm{\hat{l}}_1, \bm{\hat{l}}_2, \ldots, \bm{\hat{l}}_T\}$.   

Fig.~\ref{fig:LSTM_Curve}(a) and~\ref{fig:LSTM_Curve}(b) show the learning curves of LSTM training for Intel 5300 NIC and Nexus 5 smartphone data. Both training processes converge at a localization error around 0.3~m when the number of epochs reaches 100. However, in the real scenario, due to the random walk behavior of users, that ideal case rarely happens. Therefore, in order to approximate our method closer to practical situations, all of the reported results below are based on the case when testing points are randomly selected whose positions are considered to be a priori unknown.

\subsection{Performance Analysis}
 \begin{table}[!t]
\centering         
\caption{Average localization errors of CNN-LSTM} \label{table:AverageErr1} 
\begin{tabular}{l c c c} 
\hline           
  &  \textbf{Intel NIC} & \textbf{Nexus 5} & \\ 
Day 1 - Quiet Time (m) & 2.3 $\pm$ 1.7 &   2.8 $\pm$ 2.1 \\
Day 2 - Steady Time (m) & 2.7 $\pm$ 1.7 &   3.1 $\pm$ 1.7\\
Day 3 - Busy Time (m) & 2.6 $\pm$ 1.6 &   3.1 $\pm$ 2.2 \\
Average (m) & 2.5 $\pm$ 1.6 &   3.0 $\pm$ 2.0 \\
\hline   
\end{tabular} 
\end{table}

\begin{figure}[!t]
\centering
\includegraphics[width=0.52\textwidth]{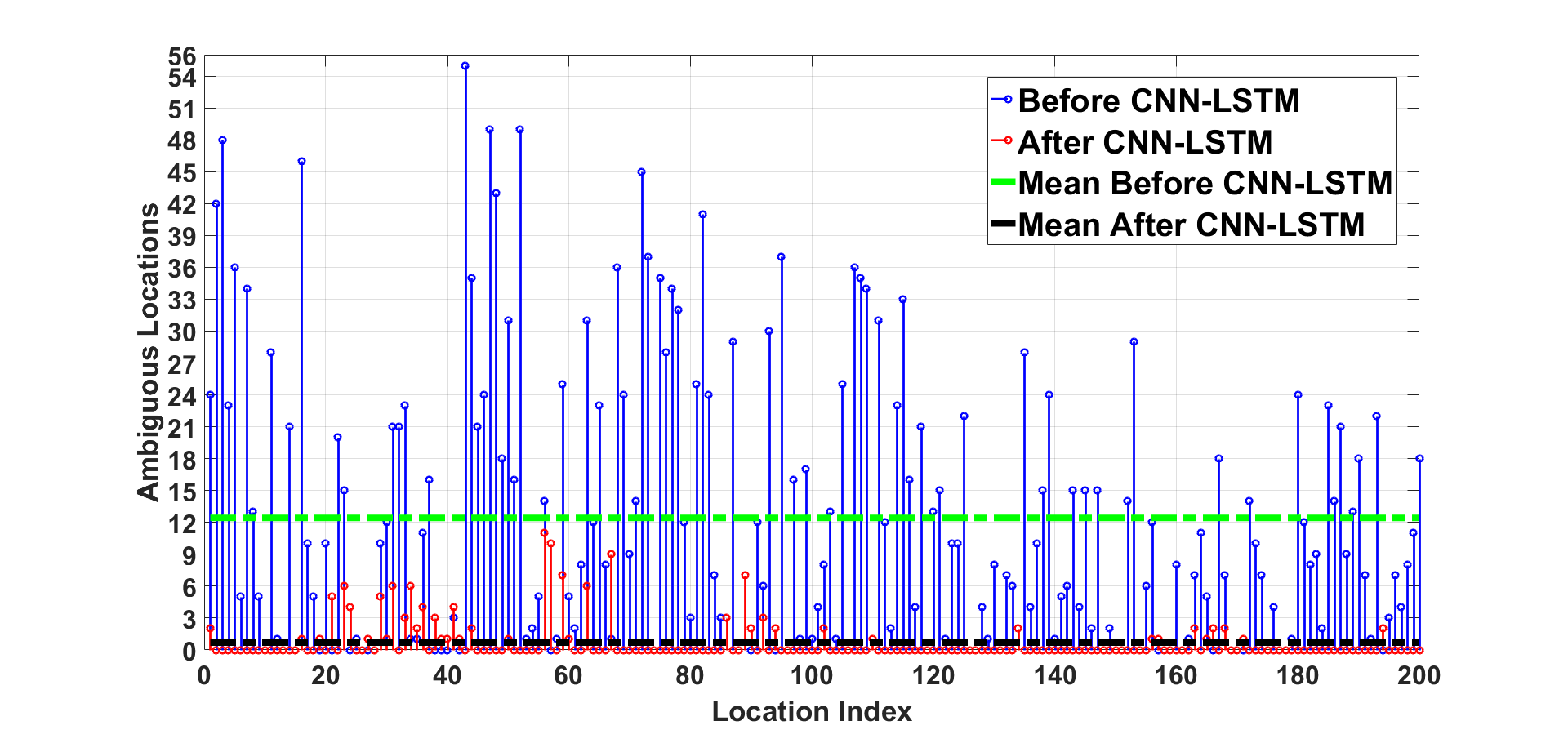}
\caption{Ambiguous locations in the database before and after CNN-LSTM process with Intel 5300 NIC dataset.}
\label{fig:AmLocs}
\end{figure}

The efficiency of CNN-LSTM layer with both space and time information is evident from the reduction of the ambiguous locations in the database before and after CNN-LSTM training presented in Fig.~\ref{fig:AmLocs}. A location $\bm{l}_{j}$ is considered ambiguous with respect to point $\bm{l}_{i}$ if they are physically far apart (beyond the experimental grid size), but their fingerprints exhibit a high Pearson correlation coefficient, exceeding the designated threshold. In our experiment, the grid size is 0.5~m. Furthermore, the correlation threshold is determined by the average correlation coefficients between $\bm{l}_{i}$ and all of its physical nearest neighbours, which is approximately 0.8 in our database. Subsequently, any non-nearest-neighbor locations with correlation coefficients exceeding this threshold are classified as ambiguous points. The fingerprints before CNN-LSTM process is the original CSI image, while the fingerprints after CNN-LSTM is the spatial output feature in $T=5$ time steps. Pearson correlation coefficient for different fingerprints is explained in details in~\cite{Minh2019}. Fig.~\ref{fig:AmLocs} shows the number of ambiguous locations among random 200 SLs of Intel 5300 NIC database. Before CNN-LSTM process, there are many locations sharing similar fingerprints (CSI images) with the average of ambiguous locations around 12. The maximum number of ambiguous locations can be up to more than 55. On the other hand, after extracting spatial and temporal information from the original CSI images, the ambiguity reduces significantly. After CNN-LSTM process, more than $80\%$ of the locations do not have any ambiguous point, which proves that their fingerprints become more distinguishable and unique. There are still few locations having ambiguity, but the maximum number of ambiguous locations are only 10 points, which is 5 times smaller than the case before the CNN-LSTM process.   
\begin{figure}[!t]
\centering
\includegraphics[width=0.52\textwidth]{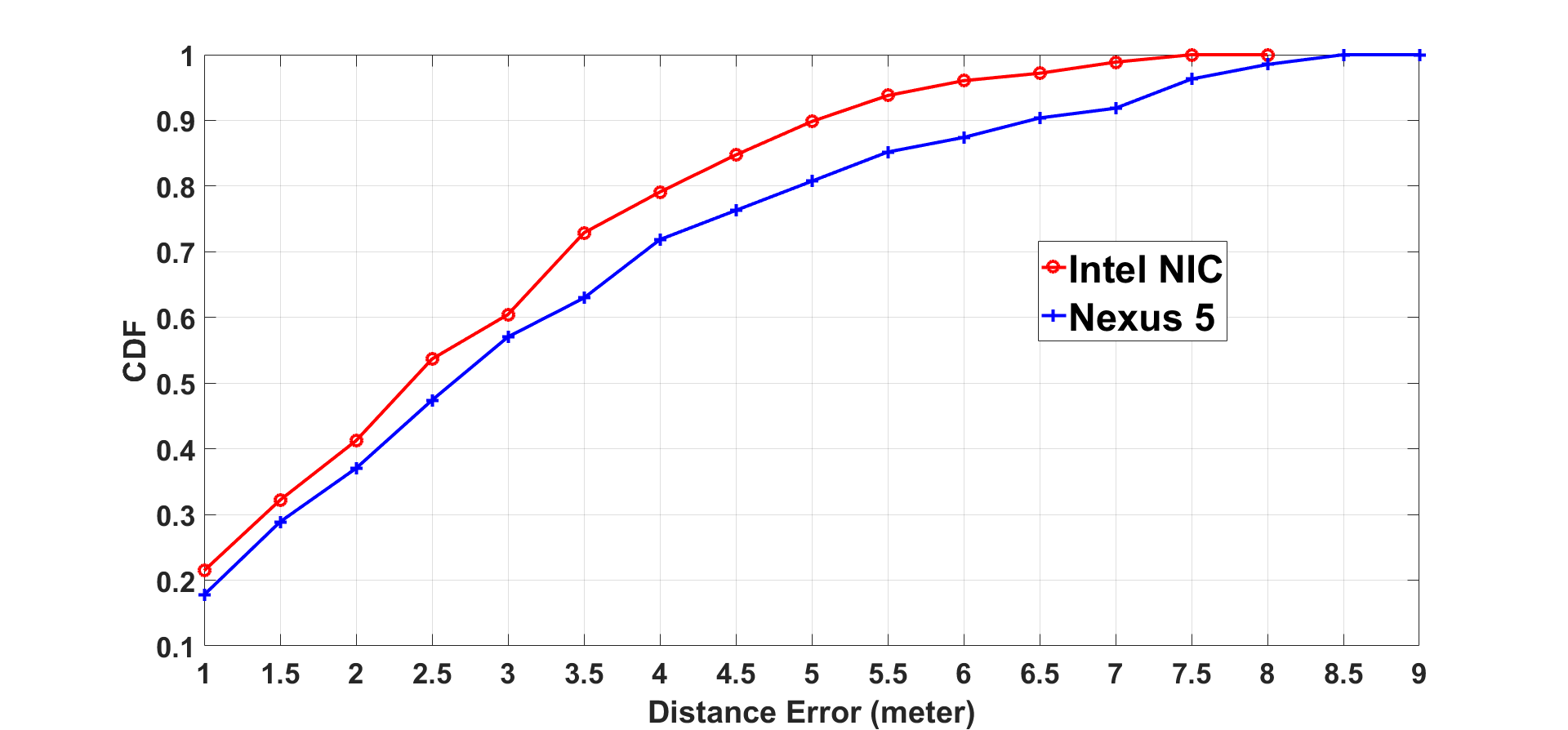}
\caption{The CDF of the localization error of the proposed CNN-LSTM with Intel NIC and Nexus 5 phone}
\label{fig:CDF1}
\end{figure}

After the training, CNN-LSTM will be utilized for realtime localization tests. Both the laptop with Intel 5300 NIC and Nexus 5 phone are mounted on the robot moving at the speed 0.6-4.0~m/s following the testing trajectory in Fig.~\ref{fig:floor_map}(a). Every $\Delta_{t}=$1-2~s, the server will receive CSI packets sending from both Intel NIC and Nexus phone, to predict the current location. Fig.~\ref{fig:CDF1} shows the CDF errors of the real testing results. Note again that all of the testing locations are placed randomly and different from all SLs. There are in total 195~collected testing points in each testing trajectory from 3 different days in different time as explained in Section~\ref{sec:experiment}. The CDF curves in Fig.~\ref{fig:CDF1} are the combination errors of all those tests. The Intel 5300 NIC has better performance than the Nexus 5 phone with $80\%$ of the localization error below 4~m compared with 5~m of the smartphone. The maximum localization error of the smartphone can be up to 9~m, while the Intel NIC's maximum localization error is only around 8~m. Table~\ref{table:AverageErr1} shows the average localization errors of both devices in 3 days of tests. The accuracy of the Intel NIC consistently outperforms the Nexus 5 phone, with the average error being around $\mathrm{2.5{\pm}1.6~m}$ and $\mathrm{3.0{\pm}2.0~m}$ respectively. The main reason that makes the difference in the performance is the number of antennae. The Intel NIC has 3 antennae, while the Nexus 5 smartphone only has a single antenna. Therefore, the CSI information containing in Intel NIC is more than in Nexus 5 phone, which leads to the more accurate localization results.          

\subsection{Results Comparison}
 \begin{table}[!t]
\centering         
\caption{Average Localization Errors - Intel 5300 NIC} \label{table:AverageErr2} 
\begin{tabular}{l c c c c c } 
\hline           
  &  \textbf{CNN-LSTM} & \textbf{BiLoc~\cite{BiLoc2017}} & \textbf{ConFi~\cite{Haochen2017}} &  \textbf{FILA~\cite{Wu2013}} \\ 
Day 1 - Quiet Time (m) & 2.3 $\pm$ 1.7 &  3.7 $\pm$ 2.6    & 6.0 $\pm$ 3.8     & 5.2 $\pm$ 2.7  \\
Day 2 - Steady Time (m)  & 2.7 $\pm$ 1.7  &  4.3 $\pm$ 3.0  & 6.2 $\pm$ 3.5    & 5.9 $\pm$ 3.2 \\
Day 3 - Busy Time (m) & 2.6 $\pm$ 1.6 &   3.7 $\pm$ 2.9  & 5.7 $\pm$ 3.2    & 5.4 $\pm$ 2.8 \\
Average (m) & 2.5 $\pm$ 1.6 &   3.9 $\pm$ 2.8  & 5.9 $\pm$ 3.5    & 5.5 $\pm$ 2.9  \\
\hline   
\end{tabular} 
\end{table}

In order to confirm the effectiveness of our CNN-LSTM model, the proposed method is compared with 3 other popular methods using CSI fingerprint including FILA~\cite{Wu2013}, ConFi~\cite{Haochen2017} and BiLoc~\cite{BiLoc2017}. All of those models are implemented under the same test environment as explained in Section~\ref{sec:experiment}. Both FILA and ConFi leverage CSI amplitude. FILA uses the probabilistic method with Bayes' rule to estimate the user's location, while ConFi adopts a CNN model with 5 layers to classify the targeted location. In contrast, BiLoc~\cite{BiLoc2017} uses both estimated AoAs and average amplitudes as input data. The mixed network, including deep autoencoder, restricted Boltzmann machine along with radial basis function, is used to get the bi-modal fingerprints. Finally, the user's location is determined by the probabilistic model with Bayes' rule. In their research, the CSI data are obtained from Intel 5300 NIC. Therefore, as a fair comparison, all of the latter figures and tables are presented with Intel 5300 NIC database results.           

\begin{figure}[!t]
\centering
\includegraphics[width=0.52\textwidth]{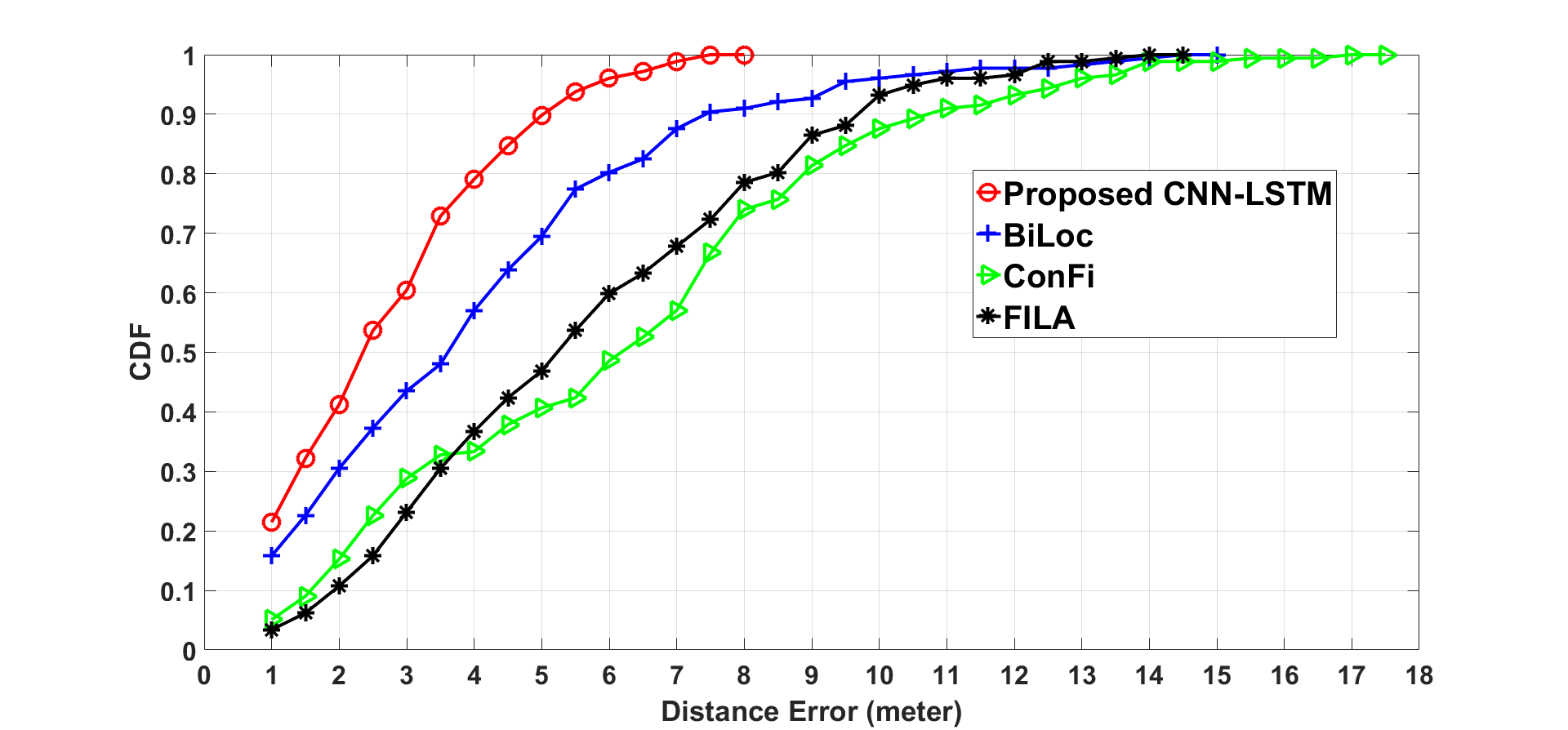}
\caption{The CDF of the localization error of the proposed CNN-LSTM and the other methods in literature.}
\label{fig:CDF2}
\end{figure}

Table~\ref{table:AverageErr2} shows the average error comparison between the proposed method and the three others. All experiments are conducted in 3 days with 195~testing locations, and the proposed model consistently has dominant performance compared with all other 3 approaches. CNN-LSTM has an average error around $\mathrm{2.5{\pm}1.6~m}$, nearly 2.5 times lower than ConFi and FILA whose errors are $\mathrm{5.9{\pm}3.5~m}$ and $\mathrm{5.5{\pm}2.9~m}$, respectively. Furthermore, the accuracy of our proposed CNN-LSTM is also $50\%$ better than BiLoc which has an average error of $\mathrm{3.9{\pm}2.8~m}$. Besides, Fig.~\ref{fig:CDF2} illustrates the CDF error comparison of all four methods. CNN-LSTM with the information of both space and time reduces the maximum error significantly, from more than 17~m of ConFi and 15~m of BiLoc to only 8~m of the proposed approach. Further, $80\%$ of the localization error of CNN-LSTM is below 4~m, which is much lower than that of BiLoc with 6~m and around 8~m of both ConFi and FILA.    

\subsection{SL Density Analysis}
\begin{figure}[!t]
     \centering 
\subfloat[\label{fig:Number_RP}]
{\includegraphics[width=0.5\textwidth]{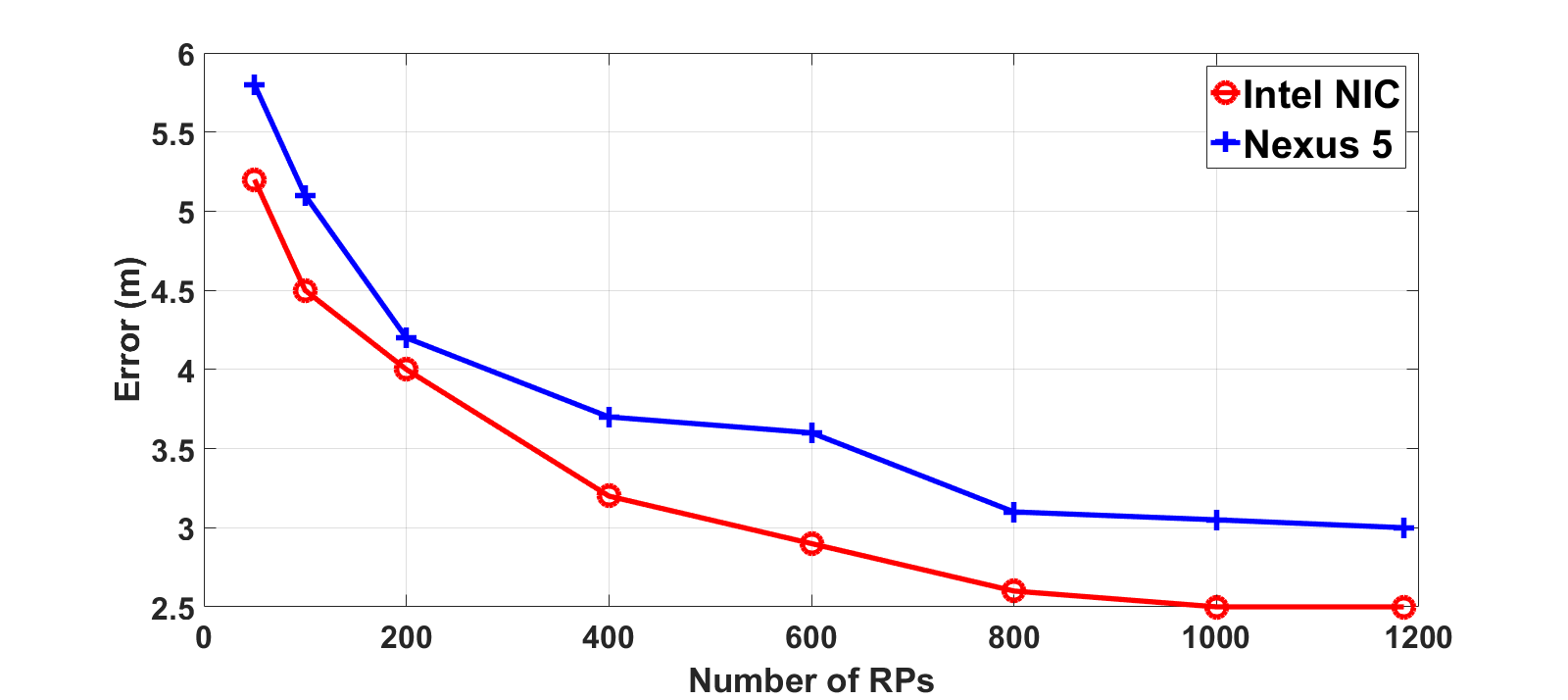}} \quad
\subfloat[\label{fig:CSI_Images}]
{\includegraphics[width=0.5\textwidth]{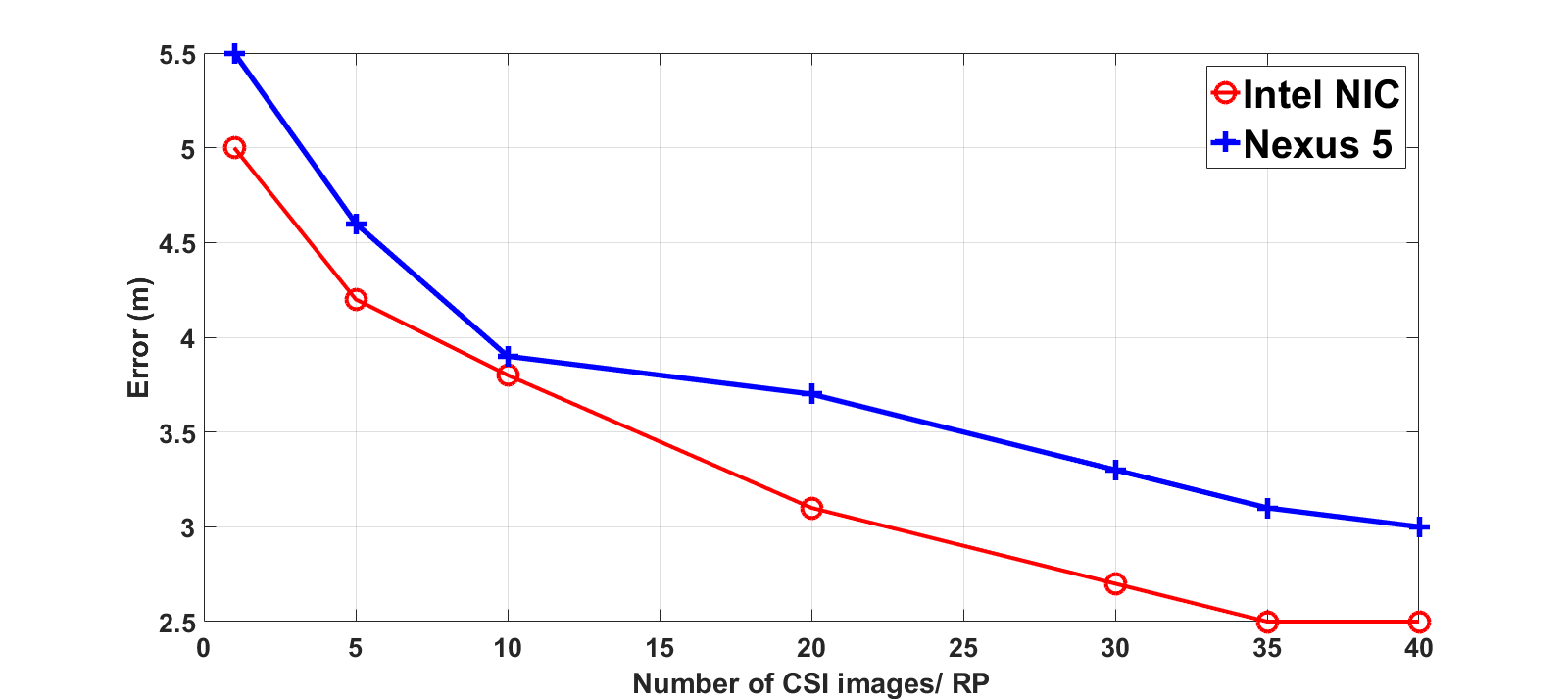}}
     \caption{Average error of the proposed network with (a) Different number of SLs. (b) Different number of CSI images per SL.}
     \label{fig:Density}
\end{figure}
In order to have a better understanding of the effects of SL density and the number of CSI images at each SL to the overall performance, the proposed CNN-LSTM network is trained with different number of input samples. Here, the reported performance is the average error of several practical tests in different days, as mentioned in Section~\ref{sec:experiment}. Fig.~\ref{fig:Density}(a) illustrates the performance of the CNN-LSTM model with different number of SLs. The number of CSI images per location are 40, while the number of SLs are varied from 50 to 1185. All chosen SLs are picked randomly from the database. For both Intel NIC and Nexus 5, the performance gets better along with the increase of the number of SLs. Starting from 5 m errors at 50 SLs, the average errors keep reducing to 2.5 m for Intel NIC and 3 m for Nexus 5 at 800 SLs. After the number of SLs increase to more than 800, the performance becomes saturated and stable. Therefore, in our test area, the proposed CNN-LSTM model has the best performance when the number of SLs reaches 800.   

On the other hand,  Fig.~\ref{fig:Density}(b) shows the performance of CNN-LSTM model with different number of CSI images per SL, where there are 1185 SLs. The number of CSI images at each SL varies from 1 to 40. Clearly, increasing the number of CSI images boosts the performance of the proposed CNN-LSTM network. When there is only 1 CSI image per SL, the average errors of Intel NIC and Nexus 5 are 5 m and 5.5 m respectively. The errors keep reducing until the number of CSI images reaches to 35. Then, after the number of CSI images per SL are more than 35, the performance of the network saturates and gets the best accuracy of 2.5 m for Intel NIC and 3 m for Nexus 5. In general, the proposed CNN-LSTM network has the best performance when we have more than 800 SLs and at least 35 CSI images per SL.    

\begin{figure}[!t]
\centering
\includegraphics[width=0.52\textwidth]{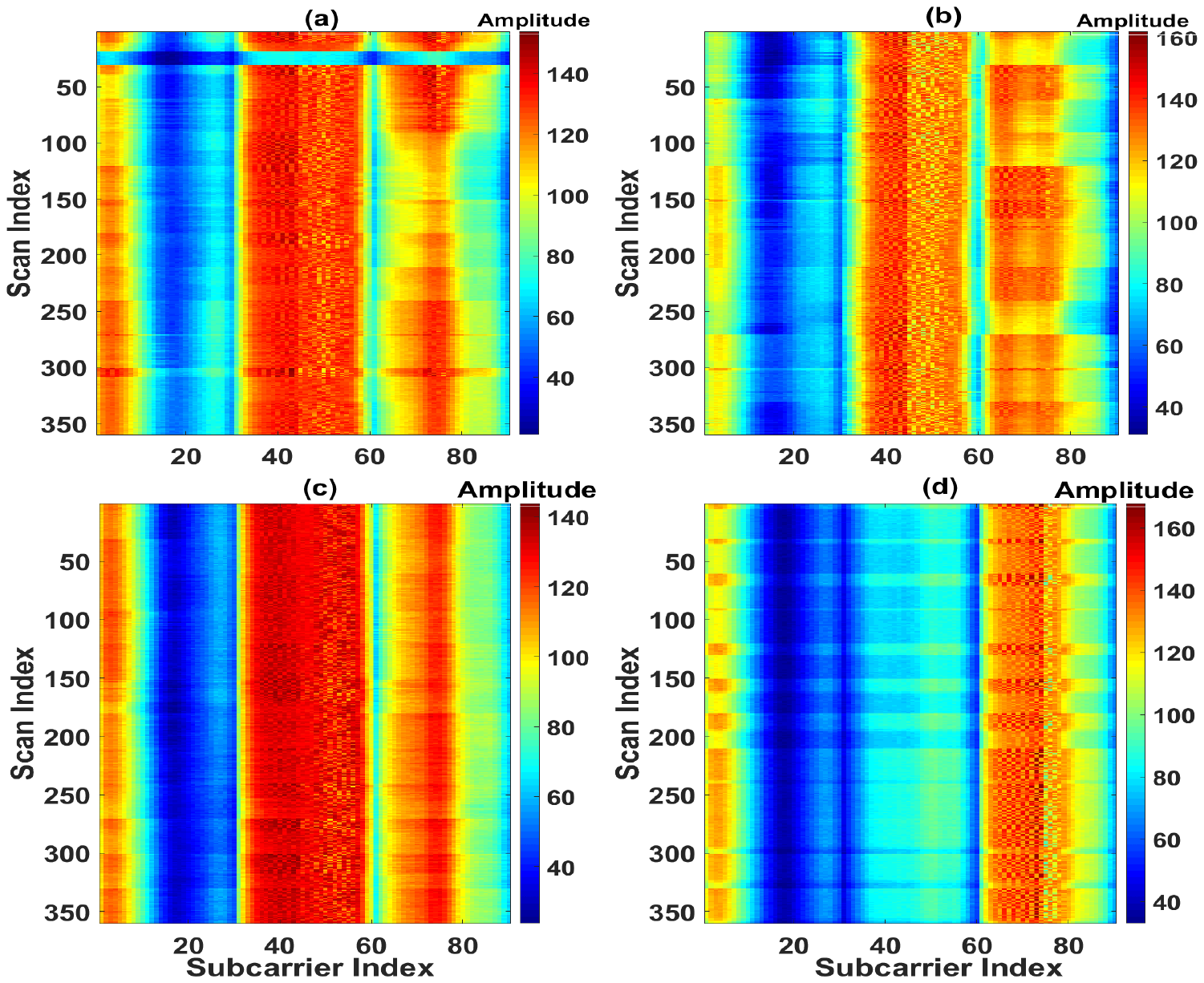}
\caption{CSI images collected from a fixed location by: (a) Robot – Orientation 1, (b) Human – Orientation 1, (c) Robot – Orientation 2, (d) Human – Orientation 2.}
\label{fig:human-robot}
\end{figure}

Regarding the conducted experiment, the CSI collected by robot could be different from what is collected by human. To quantify this difference, CSI data by both robot and human at a fixed location is collected. The experiments are measured at the same time and at the same height. There are 2 test orientations, including orientation 1: the human body does not block the WiFi signal, orientation 2: the human body blocks the WiFi signal. Note that for the robot, the phone is on the top platform, and hence the WiFi signal is never blocked by the robot. As shown in Fig. \ref{fig:human-robot} (a) and (b), in the case of no blocking for both cases, the CSI images between robot and human experiments are quite similar (85$\%$ correlation). However, in the orientation 2 with the blocking of human body, we can clearly notice the difference between 2 experimental images. The correlation coefficient between them drops to 60$\%$. Although having that difference, the data collected from the robot is believed to be reasonably close to the data collected manually from human. We applied the proposed CNN-LSTM model to both types of the collected data. The results in many different locations show that the localization accuracy has approximately 10$\%$ difference in the case of no blocking and 25 $\%$ difference in the case of blocking. Further, we can improve the accuracy by placing a dummy head or torso filled with liquid on the robot to simulate a human body.

\subsection{Apply to multiple APs scenarios}
Previously, we proved that our proposed CNN-LSTM can be fit perfectly to the environment with only one single available AP. As mentioned in \ref{sec:intro}, only a limited number of WiFi devices can provide CSI readings, for example, Intel WiFi Link 5300 MIMO NIC and Nexus 5 phone. Furthermore, in order to get continuous CSI data from a specific AP, the device needs to connect to that AP and get long preamble frame (data frame, probe request frames, etc. ~\cite{Wang2017b, Matthias2018}. Since at one time, the mobile device can only be connected to 1 AP, it can only get CSI data from that single connected AP.  Therefore, in the literature so far, there is no existing experiment about active localization which provides multiple CSI data available from multiple APs. However, in the recent research about passive indoor localization \cite{Minh2021}, Hoang \textit{et al.} claims that they use ESP32, a low-cost, low-power system on a chip microcontroller with integrated WiFi that supports to retrieve CSI information from WiFi data packets. They have up to 5 APs in the environment and at the same time can have 5 CSI data from those APs. In order to prove that our CNN-LSTM can also be applied to the scenario if we do have multiple CSI data available from multiple APs, similar steps can be done following Subsection. \ref{sec:proposed_system}. The only difference is that instead of having a single CSI images, there are 5 different CSI images are conactenated before feeding to CNN layer. Applying CNN-LSTM to the experiment with office experiment and Nexus 5, the average accuracy result of $1.5\pm 0.9$ m proves that our CNN-LSTM can also work well in multiple scenarios.

\section{Conclusions} \label{sec:conclude}
In conclusion, we have proposed a combined CNN-LSTM quantification model for accurate single router WiFi fingerprinting indoor localization. Our CNN-LSTM network extracts both spatial and temporal information of received CSI signals to determine a user's moving path. We propose a close-to-real extensive on-site experiments in a medium area with only a single available AP, with several mobile devices including mobile phone (Nexus 5) and laptop (Intel 5300 NIC) in hundreds of testing locations and in different time slots. The experimental results have consistently demonstrated the limitation of the performance of the existing works in practical scenarios and also indicated that our CNN-LSTM structure achieves an average localization error of 2.5~m with $\mathrm{80\%}$ of the errors under 4~m, which outperforms the other algorithms by at least $\mathrm{50\%}$ under the same test environment. The results of our paper also revealed the accuracy limitation of using a single router to do localization. Our proposed method and database can be used for the comparison baseline for other research. In the future research, we will utilize the proposed CNN-LSTM model in more test scenarios with different kinds of fingerprints including both RSSI and CSI to improve the performance. Some other models including hidden Markov model (HMM) and conditional random fields (CRF) are also worthy to dig deeper for the comparison. Furthermore, we will make improvements to the robot so that it can simulate a human more closely. One of these ideas, for example, is to add a mannequin to the robot.    

\bibliographystyle{IEEEtran}
\bibliography{CNN_LSTM_Ref}

\begin{IEEEbiography}[{\includegraphics[width=1in,height=1.25in,clip,keepaspectratio]{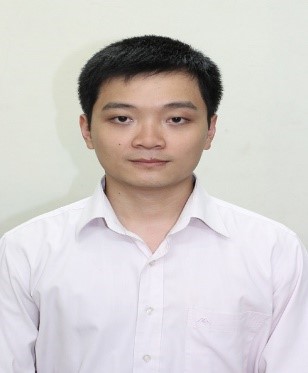}}]{Minh Tu Hoang}  received the B.Sc. degree in electronics and telecommunication engineering from the Hanoi University of Technology, Vietnam, in 2013, and Ph.D. degree with the University of Victoria, BC, Canada in 2020. He is currently working in Fortinet Canada Inc. as an embedded software developer. His research focuses on indoor localization and machine learning. From 2013 to 2016, he was a DSP Engineer with Viettel Research and Development Institute, Hanoi, Vietnam.
\end{IEEEbiography}

\begin{IEEEbiography}[{\includegraphics[width=1in,height=1.25in,clip,keepaspectratio]{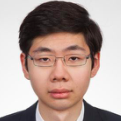}}]{Brosnan Yuen}  is currently pursuing his Ph.D in Electrical Engineering at the University of Victoria, Victoria, Canada. He is currently doing research in robotic systems, ECGs, optics, machine learning, and FPGAs.
\end{IEEEbiography}

\begin{IEEEbiography}[{\includegraphics[width=1in,height=1.25in,clip,keepaspectratio]{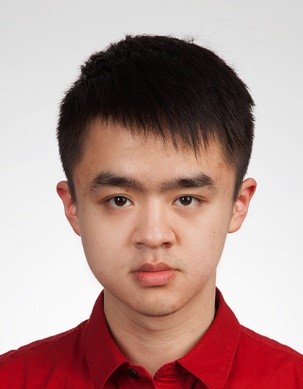}}]{Kai Ren} i received his B.Eng in Electrical and Computer Engineering from University of Victoria, 2020. He is currently pursuing a M.A.sc degree in UBC, his research direction is safe and robust control and motion planning in uncertain, dynamic environments and address applications from autonomous driving to robotics control.
\end{IEEEbiography}

\begin{IEEEbiography}[{\includegraphics[width=1in,height=1.25in,clip,keepaspectratio]{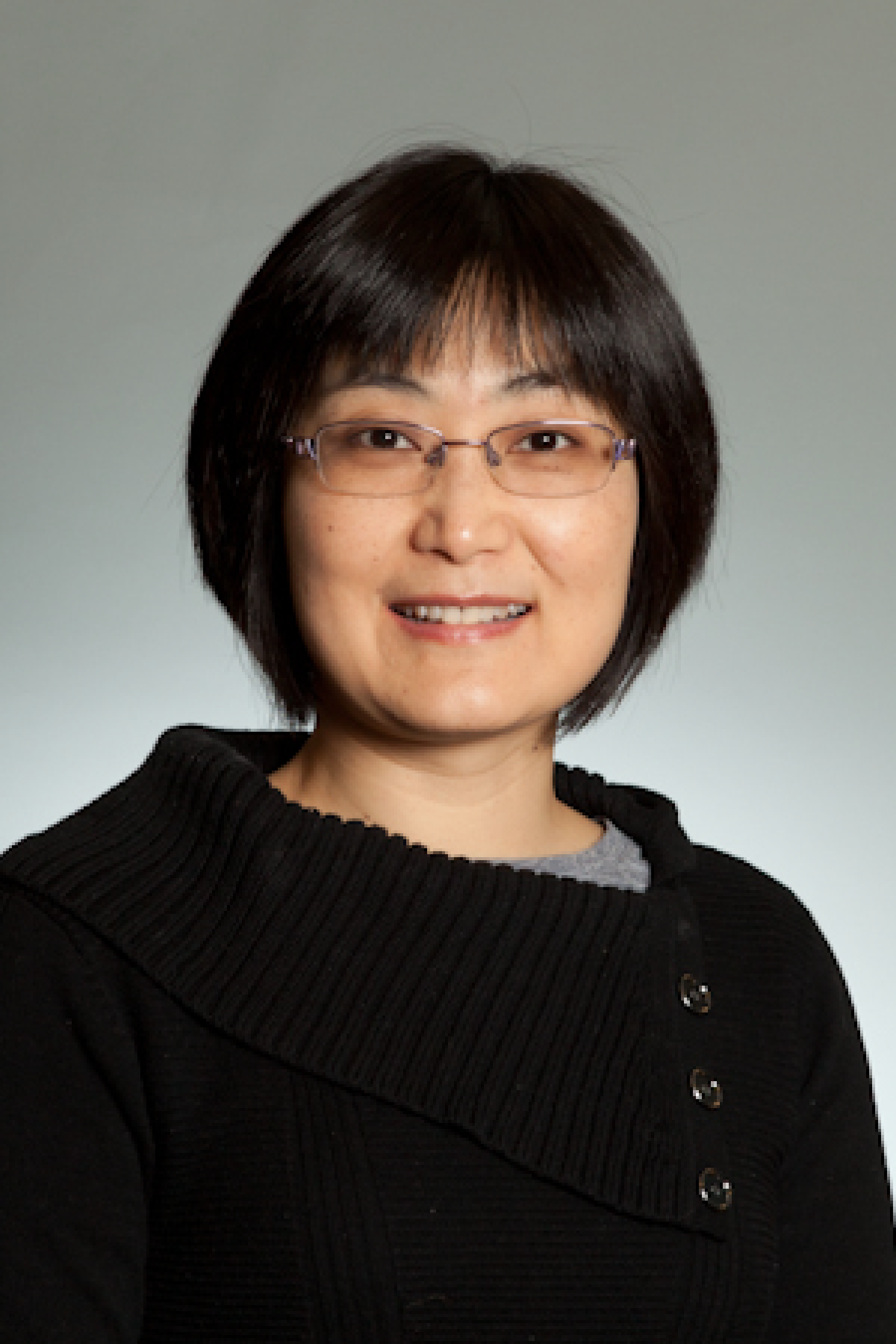}}]{Xiaodai Dong}  is currently a Professor of with the Department of Electrical and Computer Engineering, University of Victoria. She was the Canada Research Chair (Tier II) from 2005 to 2015. Dr. Dong's research interests include 5G, mmWave communications, radio propagation, Internet of Things, machine learning, localization, wireless security, e-health, smart grid, and nano-communications. She served as an Editor for the IEEE Transactions on Communications from 2001 to 2007 and the IEEE Transactions on Wireless Communications from 2009 to 2014. She is currently an Editor of the IEEE Transactions on Vehicular Technology.
\end{IEEEbiography}

\begin{IEEEbiography}[{\includegraphics[width=1in,height=1.25in,clip,keepaspectratio]{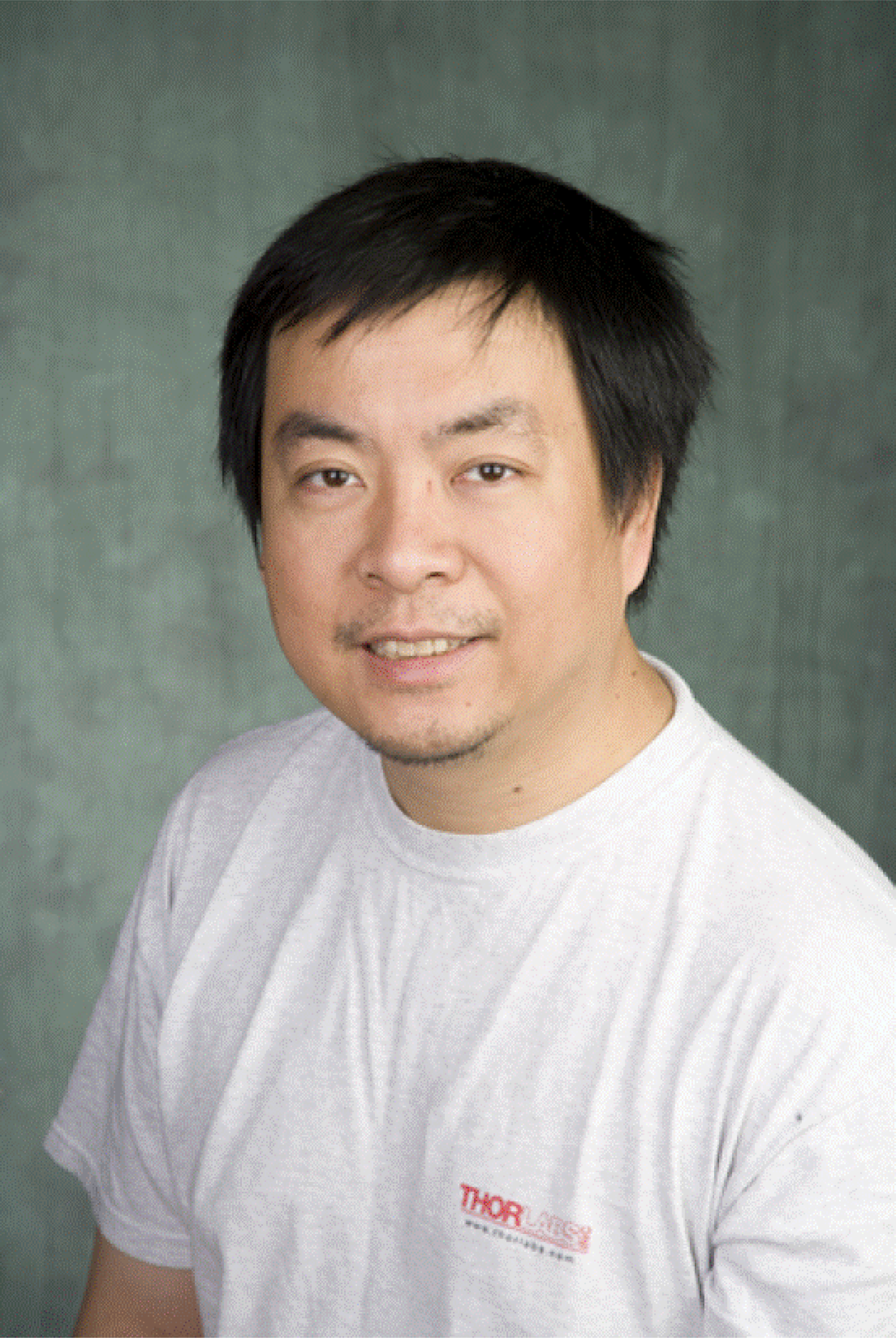}}]{Tao Lu} has worked in industry with various companies including Nortel Networks, Kymata Canada, Peleton, etc., on optical communications. Before joining the University of Victoria, he was a Post-Doctoral Fellow with the Department of Applied Physics, California Institute of Technology, from 2006 to 2008. His research interests include optical microcavities and their applications to ultra narrow linewidth laser source, and bio and nano photonics. He is currently interested in research on machine learning algorithms with applications to spectral analysis, Internet of Things, and indoor localization.
\end{IEEEbiography}

\begin{IEEEbiography}[{\includegraphics[width=1in,height=1.25in,clip,keepaspectratio]{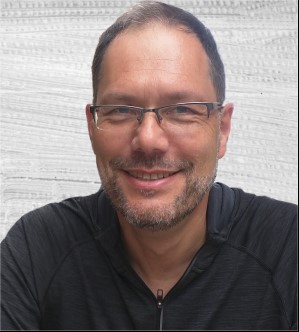}}]{Robert Westendorp } received the Dipl.Eng. degree in electrical engineering and the Ph.D. degree from the Technical University of Munich, Germany. After focusing on automation and control, he continued researching distributed operating systems for industrial measurement and automation for which he received the Ph.D. degree. His professional career led into designing hardware and software for video surveillance and intrusion detection. He is currently managing a product portfolio, including video systems and authentication servers with Fortinet Canada Inc.
\end{IEEEbiography}

\begin{IEEEbiography}[{\includegraphics[width=1in,height=1.25in,clip,keepaspectratio]{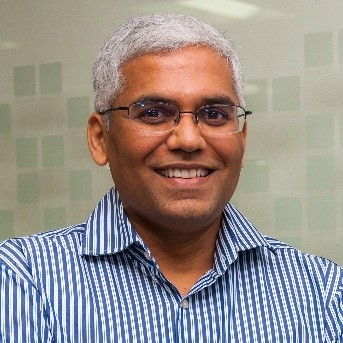}}]{Kishore Reddy Tarimala}  received his B. Tech in Electronics Communication Engineering from Jawaharlal Nehru Technological University, India. His professional career spanned designing hardware and software for world market in test, measurement instruments, consumer electronics in analog and digital video, and leading large chip design projects in semiconductors. He is currently VP of engineering for enterprise class WiFi products at Fortinet India.
\end{IEEEbiography}

\begin{IEEEbiography}[{\includegraphics[width=1in,height=1.25in,clip,keepaspectratio]{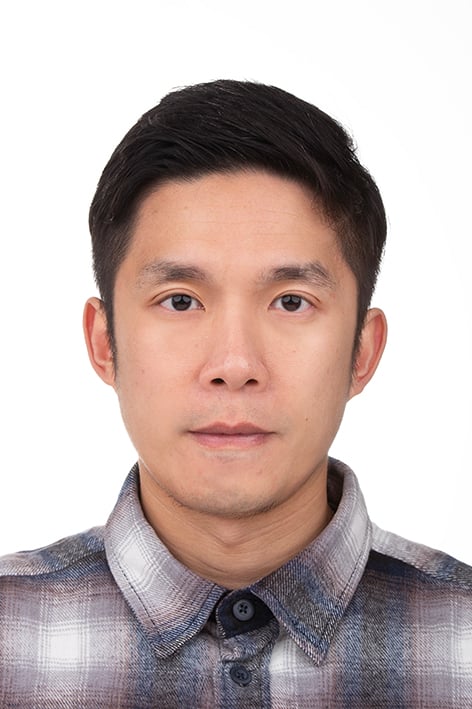}}]{Hung Le Nguyen} received the M.Sc. and B.Sc. degree in Electrical Engineering from the University of Ulsan, South Korea and Vietnam National University, Hanoi in 2018 and 2013, respectively. Since 2021, he has been working as a Ph.D. candidate in Electrical and Computer Engineering department at University of Victoria, Canada. His research interests include wireless communications and machine learning. He was a software engineer with BNF Technology Korea from 2019 to 2021 and Samsung R{\&}D Vietnam Mobile division from 2014 to 2016.
\end{IEEEbiography}

\end{document}